\documentclass[sigconf]{acmart}

\usepackage{amsmath, bbm}
\usepackage{multirow}
\usepackage{float}
\usepackage{cuted, tabularx}
\usepackage{lipsum}
\usepackage{xcolor}
\usepackage{tikz}
\usepackage{svg}
\usepackage{graphicx}
\usepackage{algorithm, algpseudocode}

\newcommand{\Vtset}{Vt$_{\text{set}}$}
\newcommand{\fio}{$\text{FiO}_{\text{2}}$}
\newcommand{\spo}{$\text{SpO}_{\text{2}}$}
\newcommand{\pao}{$\text{PaO}_{\text{2}}$}
\newcommand{\paco}{$\text{PaCO}_{\text{2}}$}
\newcommand{\pfr}{$\text{PaO}_{\text{2}}\text{FiO}_{\text{2}}\text{ Ratio}$}
\newcommand{\peep}{$\text{PEEP}$}

\AtBeginDocument{%
  }

\setcopyright{acmlicensed}
\copyrightyear{2018}
\acmYear{2018}
\acmDOI{XXXXXXX.XXXXXXX}

\acmConference[IEEE/ACM CHASE '25]{The 10th IEEE/ACM Conference on Connected Health: Applications, Systems and Engineering Technologies}{June 24--26,
  2025}{New York, NY}
\acmBooktitle{CHASE '25: The 10th IEEE/ACM Conference on Connected Health: Applications, Systems and Engineering Technologies,
 June 24--26, 2025, New York City, NY}
\acmISBN{978-1-4503-XXXX-X/18/06}

\begin{document}

\title{Matching-Based Off-Policy Evaluation for Reinforcement
Learning Applied to Mechanical Ventilation}

\author{Joo Seung Lee}
\affiliation{%
  \institution{University of California, Berkeley}
  \city{Berkeley}
  \state{CA}
  \country{USA}
}
\email{jooseung_lee@berkeley.edu}

\author{Malini Mahendra}
\affiliation{%
  \institution{University of California, San Francisco}
  \city{San Francisco}
  \state{CA}
  \country{USA}
}
\email{malini.mahendra@ucsf.edu}

\author{Anil Aswani}
\affiliation{%
  \institution{University of California, Berkeley}
  \city{Berkeley}
  \state{CA}
  \country{USA}
}
\email{aaswani@berkeley.edu}


\begin{abstract}

Mechanical ventilation is a critical life-support intervention that delivers controlled air and oxygen to a patient’s lungs, assisting or replacing spontaneous breathing. While several data-driven approaches have been proposed to optimize ventilator control strategies, they often lack interpretability and alignment with domain knowledge, hindering clinical adoption. This paper presents a methodology for interpretable reinforcement learning (RL) aimed at improving mechanical ventilation control as part of connected health systems. Using a causal, nonparametric model-based off-policy evaluation, we assess RL policies for their ability to enhance patient-specific outcomes—specifically, increasing blood oxygen levels (\spo), while avoiding aggressive ventilator settings that may cause ventilator-induced lung injuries and other complications.
Through numerical experiments on real-world ICU data from the MIMIC-III database, we demonstrate that our interpretable decision tree policy achieves performance comparable to state-of-the-art deep RL methods while outperforming standard behavior cloning approaches. The results highlight the potential of interpretable, data-driven decision support systems to improve safety and efficiency in personalized ventilation strategies, paving the way for seamless integration into connected healthcare environments.
\end{abstract}



\received{6 January 2025}

\maketitle

\section{Introduction}
Mechanical ventilation is a critical life-support intervention involving the use of a ventilator, a machine that delivers a controlled mixture of air and oxygen to the patient's lungs to facilitate proper gas exchange and to assist or replace the spontaneous breathing of patients who are unable to maintain adequate respiratory function. Respiratory failure requiring mechanical ventilation can result from various medical conditions such as infection, surgery, or trauma.

Improper administration of mechanical ventilation poses a considerable risk, potentially leading to ventilator-induced lung injuries and other complications \cite{Blauvelt2022, SR2013}. Specifically, choice of high tidal volumes is known to cause inflammation and pulmonary edema \cite{PHADF2010, SR2013}, and high fraction of inspired oxygen is known to cause hyperoxia \cite{Rachmale1887}, which all have deleterious outcomes in patients.

\subsection{Reinforcement Learning for Dynamic Treatment}

Data-driven decision-making has emerged as a powerful paradigm in healthcare, where clinicians routinely face complex dynamic treatment regimes. Reinforcement Learning (RL) methods are increasingly explored for their ability to leverage historical patient trajectories to inform sequential treatment strategies that could optimize patient outcomes. Compared to traditional static treatment guidelines, RL offers a data-centered framework for learning personalized, time-varying interventions that adapt to the evolving condition of each patient.

However, successful application of RL in healthcare requires careful consideration of several challenges. From a methodological standpoint, RL for dynamic treatment regimes involves learning policies from high-dimensional and noisy observational data. The state space consists of continuous physiological signals and clinical variables, which must be faithfully represented to avoid introducing modeling biases. Unlike controlled laboratory settings, in the clinical context we rely solely on retrospective data, making RL inherently “offline.” This offline setting must address issues such as confounding, selection bias, and distribution shifts to ensure that learned policies are both reliable and robust.

Research in RL for healthcare has shown promise across diverse clinical domains, including HIV management \cite{Yu2019, Parbhoo2017-zr}, sepsis treatment \cite{KCB18, NCLS22}, insulin dosing \cite{Oroojeni_Mohammad_Javad2019-lw, Emerson2023}, and mechanical ventilation \cite{ventai, deepvent}. By framing treatment as a sequential decision-making process, RL agents aim to optimize patient outcomes over time through a policy that suggests which therapeutic action to take given the current state of the patient. As most RL approaches in clinical care rely on offline data, recent advances in offline RL have been instrumental in defining robust learning strategies under distribution constraints.

Despite these advances, key obstacles remain. A major barrier to clinical deployment is that state-of-the-art RL policies, often realized through deep neural networks, tend to be “black boxes” with limited interpretability. Clinicians require transparency and clear rationale for suggested interventions—particularly critical for life-supporting treatments—before they can trust and adopt RL-guided strategies in practice. Similarly, commonly used Off-Policy Evaluation (OPE) methods in RL often produce high-variance or opaque estimates of expected performance, further hindering understanding and acceptance by clinicians. Improved methods are needed that not only produce accurate quantitative evaluations, but also enable intuitive, clinically relevant interpretations of how candidate policies alter patient trajectories.

\subsubsection{Reinforcement Learning for Mechanical Ventilation}

RL-based decision support lends itself well to mechanical ventilation setting well, as clinicians must continually adjust ventilator settings such as tidal volume (\Vtset) and fraction of inspired oxygen (\fio) in response to changing patient conditions. The complexity and high stakes of these decisions have motivated recent efforts to apply RL to optimize ventilation strategies \cite{ventai, deepvent, prasad2017reinforcement, Lee2023, denHengst2024}. For instance, tabular Q-learning and deep RL methods have been explored to reduce mortality and other complications, while providing automated or semi-automated guidance on setting \Vtset,\text{ }positive end-expiratory pressure (PEEP), and \fio. Although these methods demonstrate potential improvements over standard clinical practice, key challenges persist. The learned policies remain difficult to interpret due to reliance on large-scale neural networks. Moreover, the OPE methods used to validate policies often fail to provide in-depth, clinically meaningful insights about how these policies will affect patient physiology, beyond coarse metrics like mortality.

A recurring issue in these applications is the choice of state representation. Some earlier attempts discretized the state space by clustering patient measurements, aiming to derive a Markov Decision Process (MDP) with manageable complexity \cite{ventai}. However, as lung physiology and patient trajectories are inherently continuous and heterogeneous, state discretization can induce biases and lose critical granularity. Recent perspectives suggest that continuous state representations are essential for reliable modeling and policy learning, as state discretization can degrade policy performance and consistency \cite{pmlr-v235-luo24f, Liu2017}.

\subsection{Contributions and Outline}
In contrast to prior work, which often relies on black-box deep RL models that lack interpretability or fail to address clinical constraints, our paper proposes a novel approach to optimizing mechanical ventilation strategies using interpretable RL and a robust off-policy evaluation (OPE) framework. Our contributions are as follows:
\begin{enumerate}
    \item \textbf{Interpretable RL for Mechanical Ventilation:} We develop an interpretable RL method based on the Conservative Q-Improvement (CQI) algorithm \cite{cqi}. Unlike deep RL approaches, our method produces decision tree-based policies that are transparent, enabling clinicians to understand and trust the suggested ventilation strategies.
    \item \textbf{A Matching-Based Off-Policy Evaluation Method:} We propose a causal, non-parametric OPE method that generates counterfactual patient trajectories, allowing for deeper policy evaluation beyond traditional value functions. This includes clinically meaningful metrics such as overall SpO$_2$ gain and the proportion of time aggressive ventilation settings are chosen.
    \item \textbf{Clinically-Informed Reward Function:} We design a reward function that directly optimizes for SpO$_2$ improvements while explicitly penalizing aggressive ventilator settings (e.g., high tidal volumes and high inspired oxygen levels) to avoid ventilator-induced complications, which better aligns the RL objective with clinical best practices \cite{SLUTSKY19931833}.
\end{enumerate}
To demonstrate the effectiveness of our approach, we compare the performance of our interpretable RL policy (CQI) against a supervised behavior cloning baseline and a state-of-the-art deep RL policy (Conservative Q-Learning). Using both traditional OPE methods and our novel OPE framework, we evaluate the policies on real-world ICU data from the MIMIC-III database. Our results show that the interpretable CQI policy achieves competitive performance while adhering to clinical constraints and providing transparent decision-making pathways.

\section{Data}
This study leverages data from MIMIC-III, a large-scale and openly accessible critical care database \cite{Johnson2016}. We extracted patient stays involving mechanical ventilation, recording patient characteristics, lab values, vital signs, and ventilator settings at 4-hour intervals. If multiple mechanical ventilation events occurred within the same ICU stay, we selected the earliest event that lasted for more than 24 hours.

\paragraph{Inclusion Criteria}
Following criteria similar to \cite{ventai}, we included patients who:
\begin{enumerate}
    \item Were at least 18 years old at admission.
    \item Had documented 90-day mortality outcomes.
    \item Had recorded tidal volumes, ensuring the use of volume-controlled ventilation (thus excluding pressure-controlled modes).
\end{enumerate}
These criteria yielded a total of 10,739 mechanical ventilation events for analysis.

\begin{figure}
    \centering
    \includegraphics[width=0.9\linewidth]{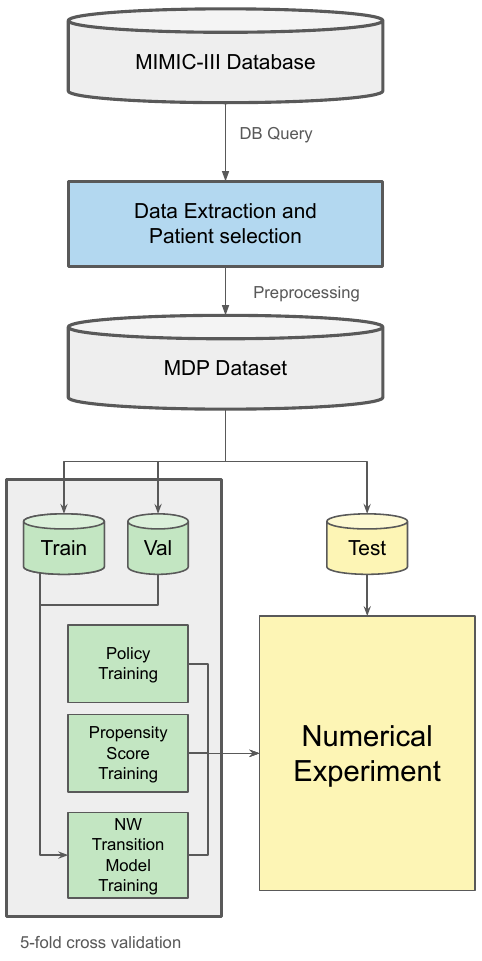}
    \caption{Overview of the pipeline for mechanical ventilation policy learning and evaluation.}
    \label{fig:enter-label}
    \Description[<short description>]{<long description>}
\end{figure}

\begin{figure*}[htb]
    \centering
    \includegraphics[width=0.9\linewidth]{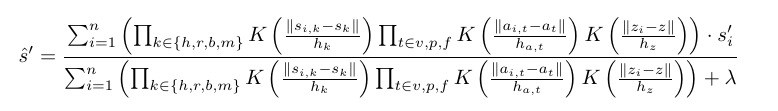}
    \caption{The Nadaraya-Watson estimator equation for modeling patient state transitions.}
    \label{fig:equation}
    \Description[<short description>]{<long description>}
\end{figure*}

\paragraph{Preprocessing Steps} Continuous features were winsorized at the 0.3th and 99.7th percentiles to mitigate the influence of extreme outliers. We truncated any ventilation event longer than 72 hours (i.e., beyond the 18th time step) to the first 18 time steps. There are four hours between each time step. Missing values were handled in a stepwise manner. Initially, we used forward and backward filling for up to five consecutive time steps. Any residual missing data were then imputed using K-nearest neighbors (KNN). Patient stays requiring more than 50\% imputation for a particular action parameter over the time course were excluded.

To provide additional clinical context, we retrieved ICD-9 codes for each patient and flagged a ventilation event as sepsis-related if the admission included at least one sepsis-associated code. Given that sepsis frequently necessitates mechanical ventilation due to acute organ failure and circulatory compromise \cite{OBRIEN20071012}, identifying these events helps differentiate respiratory failure arising from severe infections versus other causes like drug intake or post-surgical recovery.

\paragraph{Data Variables}
Each patient episode includes six time-invariant demographic and outcome-related variables (e.g., age, sex, 90-day mortality) and 20 time-dependent variables measured at each 4-hour interval. We categorize these variables as follows:
\begin{enumerate}
    \item Demographics (6): Sex, Sepsis, Weight, Age, 90 days mortality, ICU readmission\footnote{A variable indicating if the patient has prior ICU admission history}
    \item Lab Values (3): PaCO$_2$, PaO$_2$, SpO$_2$
    \item Vital Signs (11): Systolic BP, Diastolic BP, Mean BP, Glasgow Coma Scale (GCS), Heart Rate, \pfr\ (P/F Ratio), Temperature, Respiratory Rate, Spontaneous Tidal Volume, Shock index, Urine output
    \item Others (6): Total IV fluids administered, Cumulative fluid balance (CBA), SIRS, SOFA, Mean Airway Pressure (MAP), Oxygenation Index
\end{enumerate}

\section{Model}
\subsection{Markov Decision Process (MDP) Definition}
Mechanical ventilation is a critical intervention in ICU, where the clinician must continuously adjust the ventilator settings based on a patient's evolving condition. This forms a finite time-horizon sequential decision-making problem, as each adjustment to the ventilator affects future patient outcomes. To capture this, we model the mechanical ventilation control problem as a Markov decision process (MDP), represented by the tuple $(\mathcal{S}, \mathcal{A}, \mathcal{P}, r, \gamma)$, where $\mathcal{S}$ is the state space, $\mathcal{A}$ is the action space, $\mathcal{P}$ is the transition dynamics, $R$ is reward function, and $\gamma$ is discount factor.
A policy $\pi:\mathcal{S}\rightarrow \Delta(\mathcal{A})$ maps patient states to distributions over actions. 

The state-value and action-value functions under $\pi$ are defined as:
\begin{equation*}
    V_\pi(s) = \mathbb{E}_\pi\left[ \sum_{t=1}^{T}\gamma^t r(s_t,a_t,s_{t+1}) \mid s_1 = s \right]
    \end{equation*}
    and
    \begin{equation*}
Q_\pi(s, a) = \mathbb{E}_\pi\left[ \sum_{t=1}^{T}\gamma^t r(s_t,a_t,s_{t+1}) \mid s_1 = s, a_1 = a \right],
\end{equation*}
where $\mathbb{E}^\pi$ denotes expectation under $\pi$ and transition dynamics.
The goal of reinforcement learning is to find an optimal policy $\pi^*$ that maximizes the expected cumulative reward $V_\pi$. In offline reinforcement learning, the task is to find optimal policy $\pi^*$ using only a pre-gathered dataset $\mathcal{D} = \left\{\left(s^{(i)}_t, a^{(i)}_t, r^{(i)}_t, s^{(i)}_{t+1}\right)_{t=1}^{T}\right\}_{i=1}^N$, without the option of real-time interaction. This is essential for safety-critical applications like mechanical ventilation, where direct experimentation is not feasible.

\subsection{RL Problem Definition}

\paragraph{State Space}
We begin with 26 candidate features and select 15 as directly observable to the RL agent\footnote{Sepsis, Weight, Age, Heart rate, Respiratory rate, SpO$_2$, \pfr, Spontaneous tidal volume, MAP, PaCO$_2$, Systolic BP, Diastolic BP, Cumulative fluid balance, \pao, GCS}. These features provide a comprehensive view of a patient’s respiratory, hemodynamic, and mental status, while weight is included because tidal volume adjustments typically scale with patient weight.

To address potential confounding in the transition dynamics, we introduce a non-time-varying score $z$ derived from a set of 10 “type” features\footnote{SOFA, SIRS, Shock Index, Total IV fluids administered, Urine output, Mean BP, Sex, ICU Readmission, Temperature, Oxygenation Index}. These type features include composite indices like SOFA, SIRS, and shock index, which are partially composed of variables already observed by the RL agent, yet still provide important additional cues about a patient’s underlying condition. Other type variables, such as mean blood pressure and urine output, offer indirect or redundant signals and are thus grouped here to capture confounding influences not directly represented in the primary observable set. We estimate $z$ by training a logistic regression model at the initial time step to predict 90-day mortality, using these type features as inputs. Although the patient’s mortality probability might evolve in response to clinical interventions during ventilation, we treat $z$ as a proxy for the patient’s \textit{type}, a set of relatively stable traits (e.g., comorbidities, genetics) that are unlikely to change significantly within the short duration of mechanical ventilation \cite{pmid20473199}. Consequently, $z$ is held constant throughout each ventilation episode, reflecting persistent patient characteristics rather than the immediate effects of treatment. This approach is consistent with prior work on policy evaluation in healthcare \cite{pmlr-v97-oberst19a, NEURIPS2019_7c4bf50b}, and $z$ will be used later in the transition model through propensity score matching.

\begin{figure*}[htb]
    \centering
    \includegraphics[width=0.9\linewidth]{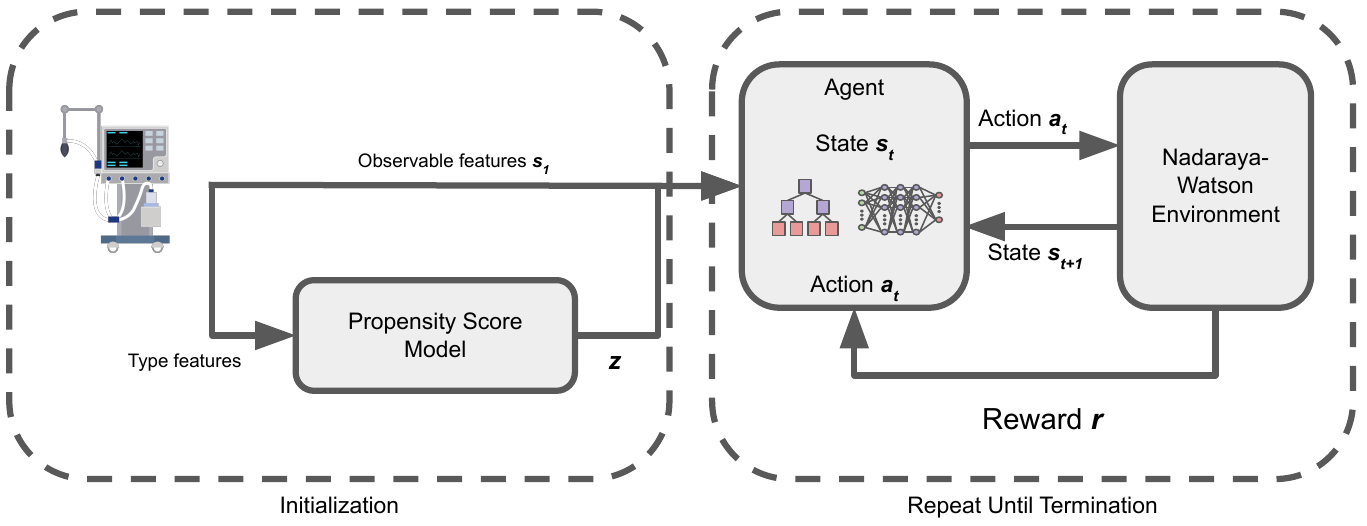}
    \caption{Overview of the matching-based off-policy evaluation framework.}
    \label{fig:dynamics}
    \Description[<short description>]{<long description>}
\end{figure*}

The final state space is therefore $\mathcal{S}\subseteq \mathbb{R}^{16}$, consisting of 15 observable patient features plus the propensity score $z$.

\paragraph{Action Space} The ventilator has the three settings, which are:
\begin{itemize}
    \item Ideal weight adjusted tidal volume (Vt$_{\text{set}}$)
    \item Positive end-expiratory pressure (PEEP)
    \item Fraction of inspired oxygen (FiO$_{2}$)
\end{itemize}

Although these settings are inherently continuous, we discretize the action space, yielding 7 levels for \Vtset, 4 for \peep, and 7 for \fio, based on a clustering analysis that revealed clinicians tend to choose from a relatively small set of commonly used ranges. This choice is further motivated by the more established theoretical foundations of off-policy evaluation in discrete action spaces \cite{pmlr-v235-luo24f}. Table \ref{fig:action-bins} in Appendix provides details of how these levels were defined and bin boundaries selected.

\paragraph{Reward}
The primary objective of operating a mechanical ventilator is to facilitate the delivery of oxygen and remove carbon dioxide from the body \cite{SLUTSKY19931833}. The effectiveness of oxygen delivery can be assessed by monitoring SpO$_2$, which represents blood oxygen levels \cite{spo23}. Although setting high values of Vt$_{\text{set}}$ and FiO$_{2}$ is correlated with immediate increases in SpO$_2$, such severe ventilator settings can increase the risk of trauma and mortality \cite{lowtidalvolume, Britos2011, Curley2016}. However, in severe cases of respiratory failure or injury, these “aggressive” settings may be necessary to maintain adequate oxygenation \cite{Curley2016, Ferguson2012}. Therefore, a RL policy that entirely avoids these aggressive values is not feasible, as it would fail to address the needs of patients in critical conditions. Nevertheless, it is undesirable to maintain patients on such settings for prolonged periods due to associated risks.

To balance these considerations, we define a reward function $r(s_t, s_{t+1}, a_t)$ that encourages moderate improvements in SpO$2$, while penalizing overly aggressive ventilator settings. The agent incurs a penalty in the form of subtracted rewards when Vt$_{\text{set}}$ is larger than 10 mmH$_2$0 or when FiO$_{2}$ is larger than 0.60, which are considered high \cite{PHADF2010, Bellani2016}.

\begin{figure*}[!htb]
    \centering
    \includegraphics[width=\textwidth]{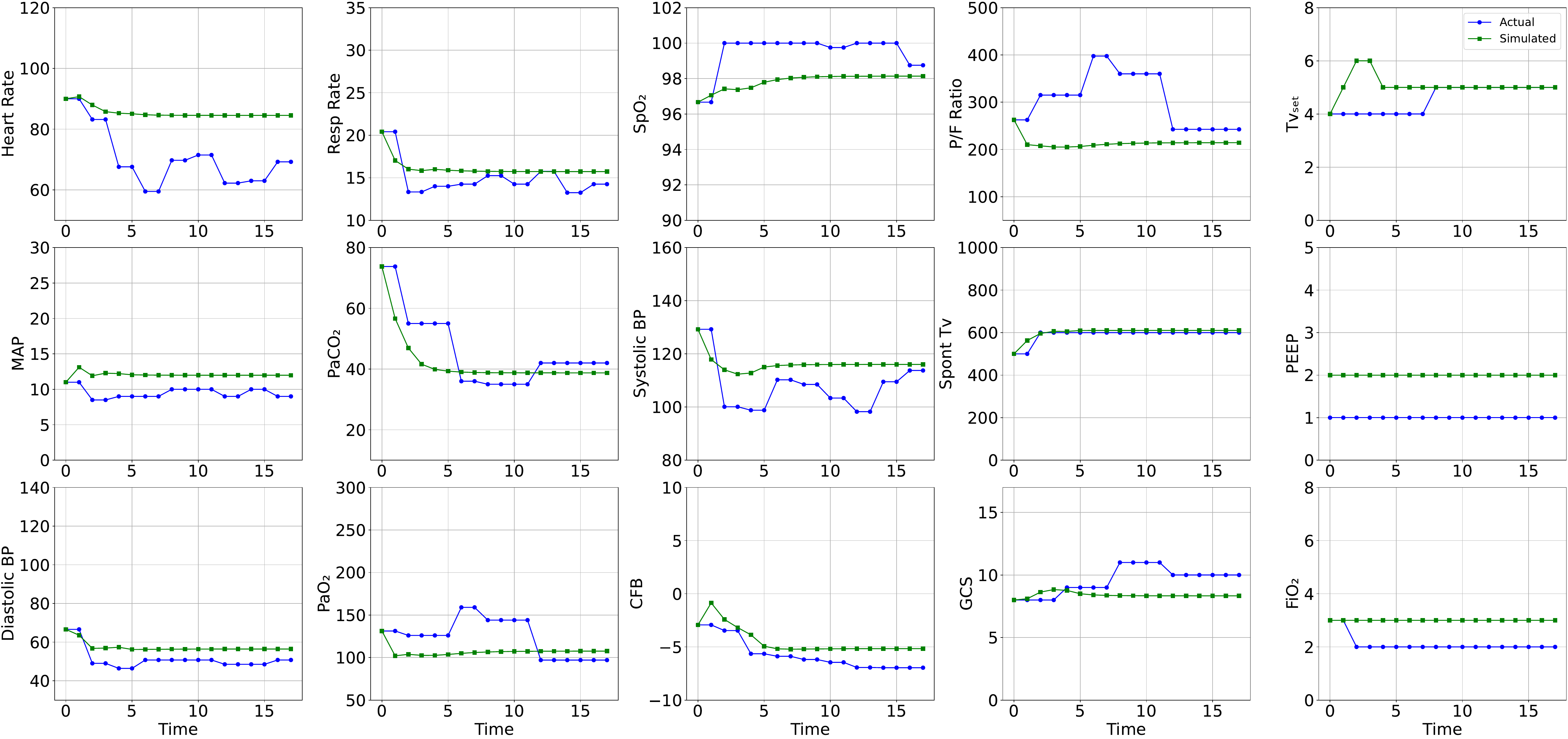}
    \caption{Sample trajectory using Nadaraya-Watson estimator transition model. This particular patient is a 50-years old female without prior ICU admission and survived for at least 90 days.}
    \label{fig:bc-sim}
    \Description[<short description>]{<long description>}
\end{figure*}

Furthermore, we design the reward function to avoid rewarding increases in SpO$_2$ when blood oxygen saturation is already at or above 95\%. This decision reflects the clinical observation that excessively high oxygen saturation can be harmful, leading to potential complications \cite{VANDENBOOM2020566}. As such, we set the reward from SpO$_2$ gain to zero is SpO$_2$ before or after the time step is above 95\%, aligning with the clinical goal of keeping SpO$_2$ within a safe range. Therefore, the final reward function $r(\cdot)$ is expressed as:
\begin{align*}
    &r^{spo}(s_t,s_{t+1}) = \begin{cases} s_{t+1}(\text{SpO}_{\text{2}}) - s_{t}(\text{SpO}_{\text{2}})\\ \quad\text{if $s_{t+1}$(SpO$_2$)$< 95$ and  $s_{t}$(SpO$_2$) $<95$} \\ 0 \qquad\text{otherwise}\end{cases}\\
    &r^{a}(a_t) = -\alpha \cdot\mathbbm{1}\{a_t(\text{Vt}_{\text{set}}) \geq 10\}
    - \beta\cdot \mathbbm{1}\{a_t(\text{FiO}_{\text{2}}) \geq 0.6\}\\
    &r(s_t, s_{t+1}, a_t) = r^{spo}(s_t,s_{t+1}) + r^a(a_t),
\end{align*}
where $\mathbbm{1}$ is the indicator function and $\alpha,\beta>0$ are constants chosen via cross-validation. The binary nature of the penalties for \Vtset and \fio\text{ }simplifies hyperparameter selection, as it avoids introducing additional parameters, making the optimization process more efficient. We use the discount factor of $\gamma=0.99$, reflecting the fact that prolonged mechanical ventilation is associated with increased risk of complications \cite{Loss2015-om}

\subsection{Policy Learning Scheme}

\paragraph{Behavior Clone (BC)} Behavior cloning is a baseline approach that treats policy learning as a supervised classification problem with the objective of imitating the historical actions taken by clinicians in the dataset. We use a random forest (a non-parametric model) to learn a behavior clone policy. Out of 196 possible combinations of actions, 164 action triplets were observed in the data. The random forest model was trained with a classification objective, which was sufficient to achieve a highly accurate behavior cloning policy. 

\paragraph{Conservative Q-Learning (CQL)} CQL algorithm \cite{cql} is designed to mitigate overestimation of Q-values in offline learning by introducing a regularization mechanism. Two regularization terms in the training objective $\mathcal{R}(\theta)$ are introduced to the standard Bellman error to minimize the overestimated Q-value at unseen actions, while maximizing it under data distribution:
\[ \mathcal{R}(\theta) = \max_\mu \mathbb{E}_{s\sim \mathcal{D},a\sim\mu(\cdot|s)}\left[Q(s,a)\right] - \mathbb{E}_{s\sim\mathcal{D},a\sim\hat{\pi}_\beta(\cdot|s)}\left[Q(s,a)\right] + \mathcal{R}(\mu)\] The d3rlpy module was used to train a CQL policy \cite{d3rlpy}.

\begin{figure*}[!t]
    \centering
\begin{tikzpicture}[
    prob/.style = {
        draw=none, 
        minimum size=0pt, 
        pos=0.3,
        xshift=-2mm
    },
    level distance=1.5cm,
    level 1/.style={sibling distance=9.5cm},
    level 2/.style={sibling distance=4.5cm},
    level 3/.style={sibling distance=2.25cm, level distance=2cm},
    every node/.style={rectangle, draw, minimum width=1.5cm, minimum height=1cm},
    leaf/.style={rectangle, draw, minimum width=1.5cm, minimum height=1cm, align=center}
]

\node {MAP $\leq$ 13.01}
    child {node {Weight $\leq$ 78.84}
        child {node {P/F Ratio $\leq$ 211.72}
            child {node[leaf] {\Vtset: 9.83 \\ \peep: 5.0 \\ \fio: 0.5} edge from parent node[prob, left] {$\leq$ 211.72}}
            child {node[leaf] {\Vtset: 9.83 \\ \peep: 5.0 \\ \fio: 0.4} edge from parent node[prob, right, xshift=4mm] {$>$ 211.72}}
            edge from parent node[prob, left] {$\leq$ 78.84}
        }
        child {node {P/F Ratio $\leq$ 211.72}
            child {node[leaf] {\Vtset: 5.62 \\ \peep: 5.0 \\ \fio: 0.5} edge from parent node[prob, left] {$\leq$ 211.72}}
            child {node[leaf] {\Vtset: 5.62 \\ \peep: 5.0 \\ \fio: 0.4} edge from parent node[prob, right, xshift=4mm] {$>$ 211.72}}
            edge from parent node[prob, right, xshift=4mm] {$>$ 78.84}
        }
        edge from parent node[prob, left, xshift=-2mm] {$\leq$ 13.01}
    }
    child {node {P/F Ratio $\leq$ 161.39}
        child {node {MAP $\leq$ 18.89}
            child {node[leaf] {\Vtset: 5.62 \\ \peep: 10.0 \\ \fio: 0.6} edge from parent node[prob, left] {$\leq$ 18.89}}
            child {node[leaf] {\Vtset: 3.94 \\ \peep: 12.5 \\ \fio: 0.6} edge from parent node[prob, right, xshift=4mm] {$>$ 18.89}}
            edge from parent node[prob, left] {$\leq$ 161.39}
        }
        child {node {Weight $\leq$ 97.49}
            child {node[leaf] {\Vtset: 5.62 \\ \peep: 10.0 \\ \fio: 0.5} edge from parent node[prob, left] {$\leq$ 97.49}}
            child {node[leaf] {\Vtset: 3.94 \\ \peep: 10.0 \\ \fio: 0.5} edge from parent node[prob, right, xshift=4mm] {$>$ 97.49}}
            edge from parent node[prob, right, xshift=4mm] {$>$ 161.39}
        }
        edge from parent node[prob, right, xshift=8mm] {$>$ 13.01}
    };

\end{tikzpicture}
    \caption{Decision tree policy with maximum depth of 3, learned from the clinicians' behavioral data.}
    \Description[<short description>]{<long description>}
\end{figure*}

\paragraph{Conservative Q-Improvement (CQI)} Conservative Q-Improvement \cite{cqi} learns a decision tree policy directly. We implemented CQI but modified to constrain the decision tree to a specified maximum depth. In contrast to CART algorithms that split a node based on information gain, the CQI algorithm splits a node based on gain of Q-values. 
Each leaf node $L$ stands for an abstract state space (intersections of half-spaces created by decision path), and the Q-value at each leaf is updated by the Q-learning update equation:
\[Q(L(s), a) \leftarrow (1-\alpha)Q(L(s), a) + \alpha(r + \gamma \max_{a'}Q(L(s_{t+1}), a')), \]
where $\alpha \in (0,1)$ is a learning rate and $L(s)$ is the leaf node corresponding to state $s$. Each branching splits the state space into two half spaces on a specific feature that leads to the Q-value gain, and the leaf node corresponds to the action with the highest Q-value in each state.

\section{Matching-Based Off-Policy Evaluation}

We use the Nadaraya-Watson estimator (NWE) \cite{Nadaraya1964OnER, watson} , a kernel-density based nonparametric regression model, to estimate next-step states for off-policy evaluation. Although neural network approaches can capture complex dynamics, they often require substantially larger datasets for stable training. By contrast, NWE is balanced between flexibility and data requirements, making a suitable choice given our dataset size. Formally, fora  current state-action pair $(s,a)$ and $z$, we model the next state as:
\[ s' = g(s, a, z) + \epsilon, \]
where $g(\cdot)$ is an unknown nonlinear function and $\epsilon$ is a noise term. To estimate $g(\cdot,\cdot,\cdot)$ from offline data ${(s_i, a_i, r_i, s'i)}_{i=1}^n$, the NWE takes a weighted average of next states $s'_i$ from similar patient transitions $(s_i, a_i)$ and propensity score $z$ in the dataset. Formally, for a given state-action pair $(s,a)$ and $z$, the NWE is defined as:
\[\widehat{g}(s, a, z) = \frac{\sum_{i=1}^n K(\|s_i - s\|/h)K(\|a_i - a\|/h)K(\|z_i - z\|/h)\cdot s'_i}{\sum_{i=1}^n K(\|s_i - s\|/h)K(\|a_i - a\|/h)K(\|z_i - z\|/h) + \lambda},\]
where $K(\cdot)$ is a kernel function,  $\lambda$ is a regularization constant, and $h$ is a bandwidth parameter.

In the context of patient state transition modeling, we apply separate bandwidths for the state variables, action variables, and the propensity score. More specifically, the 16 visible state variables are divided into four clinically oriented groups: hemodynamic, respiratory, blood gas, and miscellaneous (outlined in Appendix A.3). Each group is assigned its own bandwidth parameter ($h_{s,h}, h_{s,r}, h_{s,b}, h_{s,m}$), allowing the model to capture the distinct rates of physiological change and dependencies within each category more effectively.

We employ Epenechnikov kernel, which has finite support.
Therefore, training samples which fall outside of bandwidth for any of state, action, or propensity score are not counted towards the Nadaraya-Watson estimate. That is, given the bandwidth parameters $h_s$, $h_a$, and $h_z$, the current state $s_0$, action taken $a_0$, propensity score $z_0$, our estimate of the patient state at the next time step $\hat{s}'_0$ is given in Figure (\ref{fig:equation}).

\subsection{Generation of Simulated Trajectories}
In the medical context, unobserved confounders can influence patient transitions. To mitigate this, we incorporate a patient-specific propensity score $z$—estimated from a logistic regression model at the start of each episode—into our kernel matching. This score remains constant during the patient’s stay and effectively stratifies the population into more homogeneous subgroups. By conditioning on $z$, we help ensure that transitions are matched not only on observable current states and actions but also on persistent latent factors, yielding more reliable counterfactual estimates.

To evaluate evaluation policy $\pi_e$, we proceed as follows:

\begin{algorithm}
\caption{Generating Simulated Trajectories}\label{alg:simulate_trajectories}
\begin{algorithmic}
\Require Evaluation policy $\pi_e$, initial state $s_1$, NWE-based transition model $g(\cdot,\cdot,\cdot)$, horizon $T$
\State $s \gets s_1$
\State $z \gets LR(s_1)$
\For{$t = 1$ to $T-1$}
    \State $a_t \sim \pi_e(s_t)$
    \State $\hat{s}'_t \gets g(s_t, a_t, z)$
    \State $s_{t} \gets \hat{s}'_t$
\EndFor
\State \Return the generated trajectory $(s_1, a_1, s_2, a_2, \ldots, s_T)$
\end{algorithmic}
\end{algorithm}

As a data preparation, training dataset is prepared in a form of MDP transitions, i.e. $\{(s^i_t, a^i_t, r^i_t, s^i_{t+1}, z_i)\}$, for the transition at time $t$ for the $i^{\text{th}}$ episode. Since age, weight, and sepsis flag are assumed to not vary in a single episode, the target variable $s_{t+1}$ has the three features removed. As seen in Algorithm 1, the propensity score $z$ is computed with the initial state of the episode. Given an evaluation policy $\pi_e$, an action $a \sim \pi_e(s_1)$ is drawn, and the Nadaraya-Watson estimator transition model predicts the next state and iterates until the specified time horizon, which is set to equal the actual duration of the patient episode. This is also illustrated in Figure \ref{fig:dynamics}.

A sample trajectory rolled out on the initial state drawn from the testing dataset with the chosen hyperparameters is shown in Fig. \ref{fig:bc-sim}.

\begin{figure*}
    \centering
    \includegraphics[width=\linewidth]{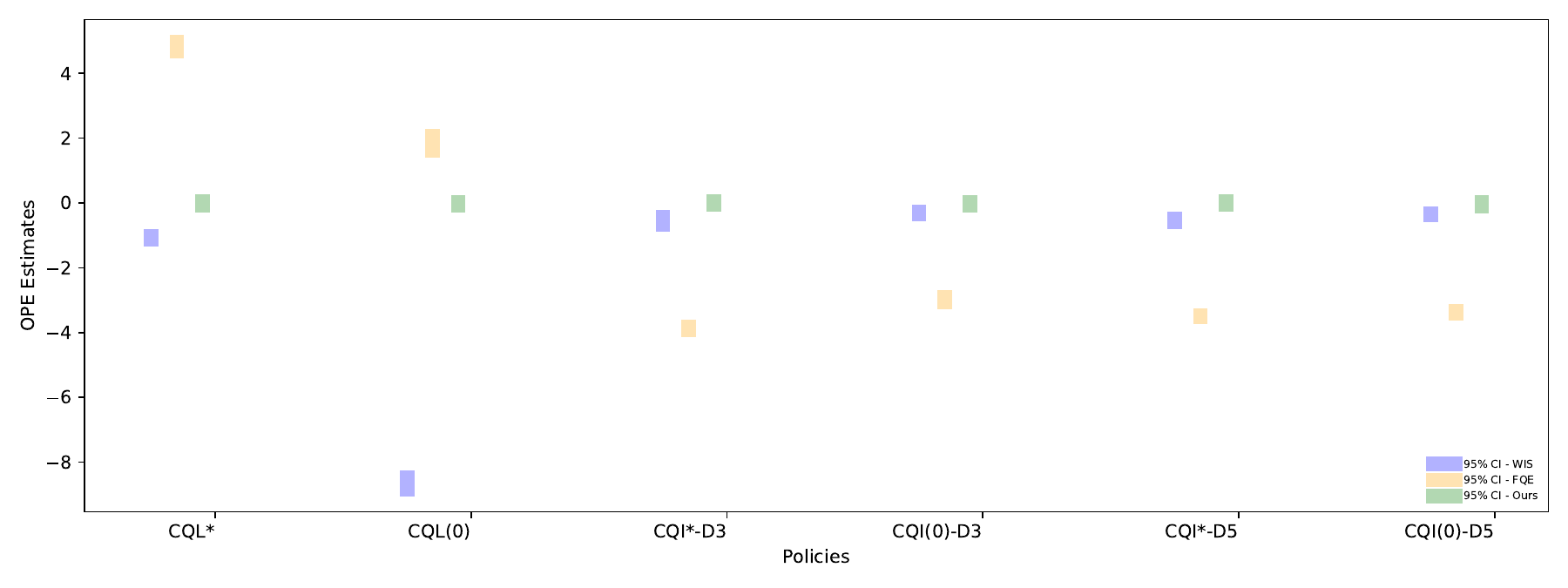}
    \includegraphics[width=\linewidth]{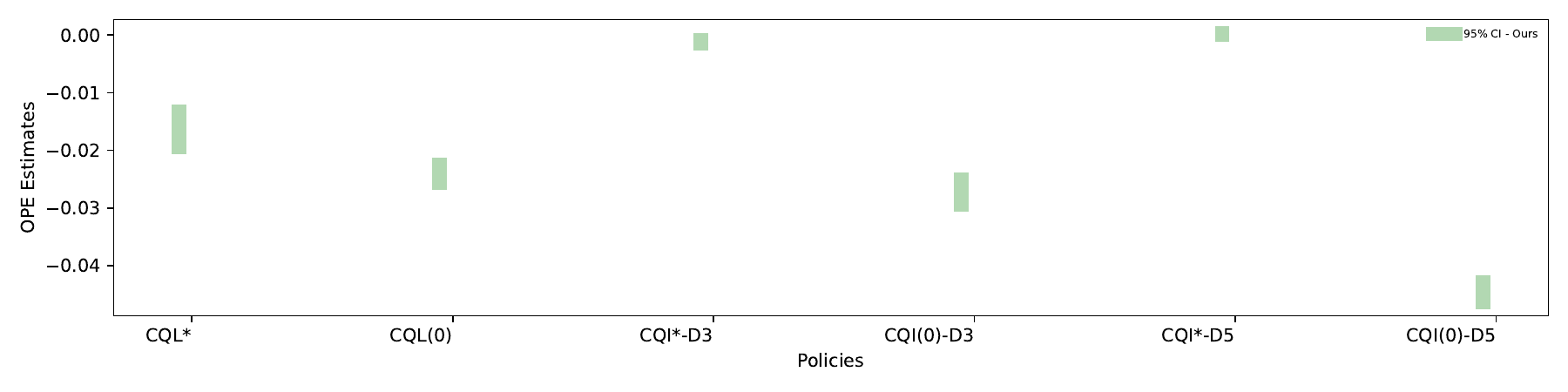}
     \caption{Comparison of OPE methods with bootstrap confidence intervals: (Upper) OPE estimates for various policies evaluated using WIS, FQE, and our nonparametric model-based OPE (Ours). (Lower) OPE estimates zoomed in to focus exclusively on the OPE estimates from our method.}
    \label{fig:ope}
    \Description[<short description>]{<long description>}
\end{figure*}

\section{Numerical Results}

\subsection{Experimental Setup}
\subsubsection{Training Scheme}

We split the mechanical ventilation events into training and testing subsets in an 80:20 ratio. Within the training set, a 5-fold cross-validation procedure was used to select hyperparameters for Behavior Cloning (BC), Conservative Q-Learning (CQL), Conservative Q-Improvement (CQI), and the Nadaraya-Watson (NW) transition model. In addition, a separate grid search was conducted to identify the penalty coefficients $\alpha$ and $\beta$ for discouraging aggressive ventilator settings.

A behavior cloning policy was trained using a random forest to replicate clinician-chosen discretized actions. Its final testing mean absolute errors (MAE) in predicting \Vtset, \peep, \fio\text{ }were $0.383$, $0.755$, and $0.825$, respectively. Concurrently, a logistic regression model achieved a testing ROC-AUC of $0.658$ for estimating each patient’s propensity score, which remained fixed throughout their ventilation episode. This logistic regression score was incorporated into our off-policy evaluation to address latent confounding.
For our transition model, we employed the Nadaraya-Watson (NW) estimator with separate bandwidth parameters for state features, action features, and the propensity score. Using an Epanechnikov kernel and cross-validation, the selected bandwidths ($h_{s,h}, h_{s,r}$, $h_{s,b}$, $h_{s,m}$, $h_a$, $h_z$) were ($3.04$, $2.8$, $2.53$, $2.03$, $1.0$, $1.5$), and the regularization constant $\lambda$ was $10^{-3}$. Under the BC policy, the tuned NW model attained a mean absolute error $1.56$ when predicting \spo\text{ }time step by time step, and an AUC-ROC of $0.716$ for distinguishing whether \spo\text{ }exceeded 94\%.
A separate grid search over the penalty coefficients $\alpha, \beta \in [\frac{3}{2^0}, \frac{3}{2^1}, \frac{3}{2^2}, \frac{3}{2^3}, \frac{3}{2^4}]$ and the CQL conservativeness parameter $\alpha_{CQL} \in [4, 2, 1, 0.5, 0.25]$ was carried out to balance maximizing \spo\text{ }improvements and minimizing aggressive ventilator settings. Aggressive actions were defined as \Vtset$\geq 6$ or \fio$\geq 4$, corresponding to the higher action indices in our discretized space (Appendix B). The optimal values were $\alpha^* = 0.375, \beta^* = 0.75$, and $\alpha_{CQL} = 0.25$. With these settings, the CQL policy achieved an average increase of \spo\text{ }of $0.070$, with \Vtset\text{ }deemed aggressive in only $1.13\%$ of time steps and \fio\text{ }aggressive in $3.46\%$ of time steps.

\begin{figure}[!ht]
    \centering
    \includegraphics[width=240pt]{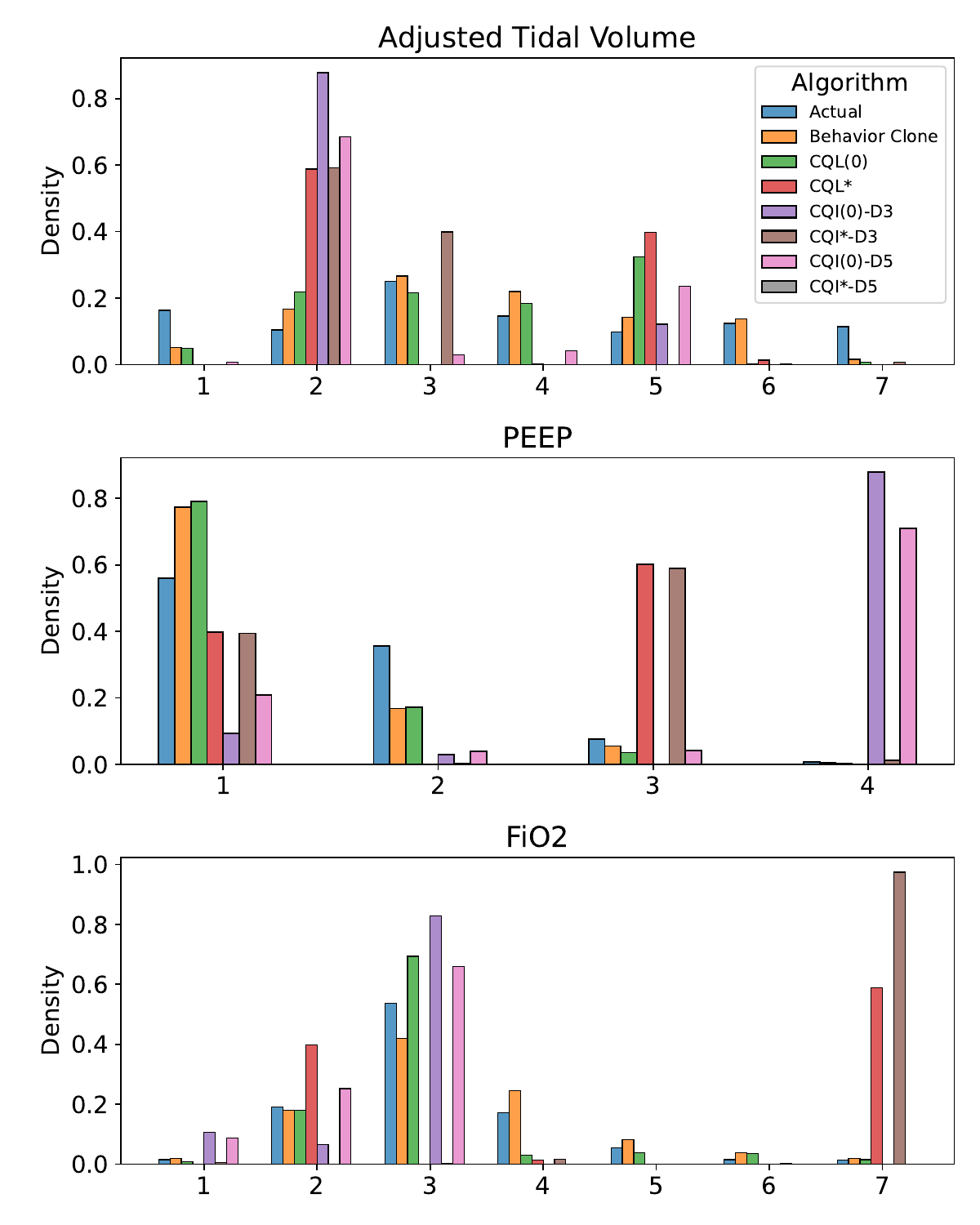}
    \label{fig:actiondist}
    \caption{Distribution of actions across different RL \& BC policies. The x-axis refers to action bins introduced in Fig. \ref{fig:action-bins}.}
\end{figure}

\begin{figure*}[!t]
    \centering
\begin{tikzpicture}[
    prob/.style = {
        draw=none, 
        minimum size=0pt, 
        pos=0.3,
        xshift=-2mm
    },
    level distance=1.5cm,
    level 1/.style={sibling distance=9.5cm},
    level 2/.style={sibling distance=4.5cm},
    level 3/.style={sibling distance=2.25cm, level distance=2cm},
    every node/.style={rectangle, draw, minimum width=1.5cm, minimum height=1cm},
    leaf/.style={rectangle, draw, minimum width=1.5cm, minimum height=1cm, align=center}
]
\node {PaCO$_2$}
    child {node {Weight}
        child {node {P/F Ratio}
            child {node[leaf] {\Vtset: 7.1\\ \peep: 5.0 \\ \fio: 0.5} edge from parent node[prob, left] {$<$ 224.23}}
            child {node[leaf] {\Vtset: 5.98\\ \peep: 5.0 \\ \fio: 0.4} edge from parent node[prob, right, xshift=4mm] {$\geq$ 224.23}}
            edge from parent node[prob, left] {$<$ 104.88}
        }
        child {node {P/F Ratio}
            child {node[leaf] {\Vtset: 7.1\\ \peep: 5.0 \\ \fio: 0.5} edge from parent node[prob, left] {$<$ 224.23}}
            child {node[leaf] {\Vtset: 4.8\\ \peep: 5.0 \\ \fio: 0.4} edge from parent node[prob, right, xshift=4mm] {$\geq$ 224.23}}
            edge from parent node[prob, right, xshift=4mm] {$\geq$ 104.88}
        }
        edge from parent node[prob, left, xshift=-2mm] {$<$ 19.54}
    }
    child {node {PaCO$_2$}
        child {node {Mean Airway Pressure}
            child {node[leaf] {\Vtset: 3.39\\ \peep: 5.0 \\ \fio: 0.4} edge from parent node[prob, left] {$<$ 3.55}}
            child {node[leaf] {\Vtset: 8.31\\ \peep: 12.5 \\ \fio: 0.5} edge from parent node[prob, right, xshift=4mm] {$\geq$ 3.55}}
            edge from parent node[prob, left] {$<$ 92.13}
        }
        child {node {SysBP}
            child {node[leaf] {\Vtset: 7.1\\ \peep: 5.0 \\ \fio: 0.5} edge from parent node[prob, left] {$<$ 96.19}}
            child {node[leaf] {\Vtset: 4.8\\ \peep: 10.0 \\ \fio: 0.5} edge from parent node[prob, right, xshift=4mm] {$\geq$ 96.19}}
            edge from parent node[prob, right, xshift=4mm] {$\geq$ 92.13}
        }
        edge from parent node[prob, right, xshift=8mm] {$\geq$ 19.54}
    };
\end{tikzpicture}
    \caption{Left-child view of the decision tree policy trained by CQI algorithm with the maximum depth of 4 and the penalty coefficients $(\alpha^*, \beta^*)$, i.e. CQI$^*$-D$4$. The first splitting point corresponds to \spo $< 97.6$. The units used for \Vtset\text{ }and \peep\text{ }are each mL/kg and cm H$_2$O.}
    \label{fig:dt-cqi}
    \Description[<short description>]{<long description>}
\end{figure*}

\subsubsection{Off Policy Evaluation}
Since direct real-time experimentation with new policies is infeasible, we rely on OPE to estimate each policy's performance. OPE leverages retrospective data to approximate the expected returns of a policy different from the one that generated the dataset, circumventing safety risks. We considered three widely used OPE methods:
\begin{enumerate}
    \item \textit{Importance Sampling} \cite{precup_is} Importance sampling reweighs observed trajectories based on the ratio of the evaluation policy to the behavior policy at each step.
    We use per-step weighted IS estimator (PSWIS), which is given by:
    \begin{align*}
        \hat{V}_{\text{step-WIS}} = \frac{1}{n}\sum_{i=1}^n \sum_{t=1}^T \gamma^t\frac{\rho^{(i)}_{1:t}}{w_t} r^{(i)}_t.
    \end{align*}
    where $\rho^{(i)}_{1:t} =  \prod_{s=1}^t \frac{\pi_e(a^{(i)}_t|s^{(i)}_t)}{\pi_b(a^{(i)}_t|s^{(i)}_t)}$ and $w_t = \sum_{i=1}^{|\mathcal{D}|}\frac{\rho^{(i)}_{1:t}}{|\mathcal{D}|}$.
    \item \textit{Fitted Q-Evaluation} Fitted Q-Evaluation \cite{fqe} is a regression-based direct method to perform OPE. It iteratively improves the estimate of the Q-function by computing the targets $y_i = r_t + \gamma Q_{k-1}(s_{t+1}, \pi_e(s_{t+1}))$ and solving the supervised problem $Q_k = \arg\min_{f\in\mathcal{F}}\frac{1}{n}\sum_{i=1}^n (f(s_t,a_t) - y_t)^2$, where $\mathcal{F}$ is a learnable function class, such as the set of neural networks.
\end{enumerate}

\subsection{Results}

Let $\pi_{BC}$ denote the behavior clone policy, $\pi_{CQL(0,0)}$ and $\pi_{CQL^*}$ each refer to CQL policy with $\alpha=\beta=0$ and $\alpha^*, \beta^*$, and $\pi_{CQI(0,0)-Dn}$ and $\pi_{CQI^*-Dn}$ refer to CQI policy with maximum depth of $n$, each with $\alpha=\beta=0$ and $\alpha^*, \beta^*$.

\subsubsection{Comparison of various OPE methods}
We empirically compare the performances of various OPE methods in Figure \ref{fig:ope}. The results reveal several key trends and distinctions across the evaluated methods: WIS, FQE, and our proposed nonparametric model-based OPE method ("Ours").

FQE consistently considers policies from the CQL family to perform better than those from the CQI family. This observation aligns with our intuition that more complex models, such as those enabled by deep reinforcement learning, achieve better outcomes by leveraging their capacity to capture nuanced patient trajectories. In contrast, WIS identifies CQL(0) as particularly poor, suggesting that this method is sensitive to the aggressive policies characteristic of unpenalized algorithms.

Our method tends to favor policies trained with penalties (e.g., $\pi_{CQL^*}$, $\pi_{CQI^*-D5}$, $\pi_{CQI^*-D5}$) over those without penalties. This trend is consistent across all policy pairs. A plausible interpretation is that more aggressive policies, which avoid penalization, tend to perform worse under our nonparametric OPE framework. This may reflect an inherent property of the method: its ability to account for penalties imposed on aggressive action choices, which are more likely with unpenalized policies. Notably, this alignment with clinical intuition underscores the clinical relevance of our OPE in prioritizing cautious policies that balance performance with patient safety.

The comparison also illustrates the balance between complexity and interpretability. While deep RL methods such as $\pi_{CQL^*}$ achieve strong performance, our method’s preference for interpretable policies like $\pi_{CQI^*-D5}$ and $\pi_{CQI^*-D5}$, when penalties are applied, suggests that interpretable models can remain competitive when designed with appropriate constraints. This balance is critical for clinical adoption, where the transparency of decision-making is as vital as quantitative performance metrics.

\begin{figure}[!ht]
    \centering
    \footnotesize
    \begin{tabular}{|l|c|c|c|c|}
        \hline
        \textbf{Algorithm} & \textbf{\% \spo $> 95$} & \textbf{$\overline{\Delta(\text{\spo})}$} & \textbf{\% \Vtset} & \textbf{\% \fio} \\
        \hline
        BC & 98.10 & -0.47 & 24.19 & 25.57 \\
        CQL(0,0) & 98.99 & 0.04 & 15.63 & 37.68 \\
        CQL$^*$ & 99.87 & 0.63 & 0.76 & 11.75 \\
        CQI(0,0)-D3 & 99.75 & -0.32 & 1.22 & 60.86 \\
        CQI$^*$-D3 & 99.03 & -1.19 & 0.00 & 0.00 \\
        CQI(0,0)-D5 & 100 & N/A & 0.86 & 99.25 \\
        CQI$^*$-D5 & 99.70 & -1.87 & 0.00 & 0.00 \\
        \hline
    \end{tabular}
    \caption{Comparisons of the different RL \& BC policies. \% \spo $> 95$ indicates the proportion of episodes that ended with terminal \spo\space{ }of 95 or above. $\overline{\Delta(\text{\spo})}$ indicates the average increase in \spo\text{ }per event, only among those whose terminal \spo\text{ }less than 95. Proportion of patients starting with \spo $ \leq 95\%$ is 13\%} 
    \label{fig:performances}
    \Description[<short description>]{<long description>}
\end{figure}

\subsubsection{Distribution of Action Choices and Performance Comparison}
The distribution of actions selected by different policies reveals significant differences in their behavior and alignment with clinical goals. As shown in Figure 8, policies trained with penalties on aggressive actions demonstrate a marked reduction in the frequency of such actions compared to both the behavior cloning policy and unregularized policies. Specifically, the CQI policies constrained by positive penalties entirely avoid aggressive actions, as evidenced by 0\% aggressiveness in both tidal volume (\Vtset $\geq 10$ mL/kg) and fraction of inspired oxygen (\fio $\geq 0.6$). This outcome reflects the structural constraints imposed by the limited number of decision tree leaves and the explicit penalty in the reward function, which collectively encourage the selection of non-aggressive actions across all leaves.

In contrast, CQL policies, which leverage a more flexible neural network architecture, achieve a balance between reducing aggressive actions and improving \spo. Among these, the regularized $\pi_{CQL^*}$ stands out, achieving the lowest frequency of aggressive actions while maintaining a positive $\Delta$(\spo) for patients whose terminal \spo\text{ }falls below 95\%. This outcome highlights the ability of $\pi_{CQL^*}$ to optimize oxygenation without relying excessively on high-risk settings. However, the unregularized $\pi_{CQL(0,0)}$, while maintaining a positive $\Delta$(\spo), exhibits higher aggressiveness in \fio, with 37.68\% of time steps classified as aggressive. This trade-off underscores the importance of regularization in achieving a clinically appropriate balance between performance and safety.

The two policies $\pi_{CQI^*-D5}$ and $\pi_{CQI^*-D5}$, while interpretable due to their decision tree structure, exhibit certain limitations. Both policies achieve 0\% aggressiveness due to the combined effect of their shallow tree depth and the reward penalties, which effectively constrain their action space. However, this strict avoidance of aggressive actions comes at the expense of performance in improving \spo\text{ }for critical patients, as evidenced by their negative $\Delta$(\spo). The negative values indicate that these policies are less effective at enhancing oxygenation in patients who require aggressive interventions.

Interestingly, $\pi_{CQI(0,0)-D5}$ achieves 100\% of episodes terminating with \spo\text{ }$> 95\%$, the highest among all policies. However, this comes at the cost of relying on a broader range of actions, including those that may not align with the penalty structure used in $\pi_{CQI^*}$. By contrast, $\pi_{CQI^*-D5}$ achieves comparable results (99.70\%) while strictly adhering to the non-aggressiveness constraint, demonstrating that interpretability and safety can coexist, albeit with some trade-offs in $\Delta$(\spo).

\paragraph{The Cost of Interpretability}

The progression from CQI-D3 to CQI-D5 to CQL reflects a fundamental trade-off in RL policy design: increasing model complexity enhances performance but reduces interpretability. CQI-D3, the shallowest decision tree, provides simple, threshold-based rules but struggles with complex scenarios, achieving a negative $\Delta$(\spo). Its strict avoidance of aggressive ventilator settings ensures safety but sacrifices performance in critical cases requiring aggressive interventions.

Increasing the tree depth to CQI-D5 marginally improves performance and retains interpretability relative to more complex models. CQI-D5 achieves nearly perfect \% \spo $> 95$ outcomes without aggressive actions, but has severe aggressiveness in \fio\text{ }strategy. Also, $\pi_{CQI^*-D5}$ has the worst outcome in $\Delta$(\spo).

The deep neural network policy (CQL) excels in $\Delta$(\spo) and minimizing aggressive actions due to its ability to model complex, nonlinear relationships. However, this performance gain comes at the cost of transparency, as its decisions are not readily explainable.

\subsubsection{Decision Tree}
The decision tree presented in Figure \ref{fig:dt-cqi} represents a distilled policy learned via reinforcement learning (RL) from mechanical ventilation data documented in the electronic health record, guided by a custom reward structure. While some decision level splits were associated with clinical reasoning typically used at the bedside, many decision level splits were based on data demarcations that are not consistent with expected measures of sepsis physiology.

A striking observation is that the policy heavily relies on \paco\text{ }as its primary decision node, subdividing patients into those with very low \paco\text{ }(<19.5 mmHg), moderate \paco\text{ }(19.5–92.1 mmHg), and very high \paco\text{ }($\geq 92.1$ mmHg). 
WHile titrating the ventilator's respiratory rate and \Vtset to goal \paco levels is an important part of mechanical ventilator management, clinicians typically titrate \paco in sepsis management to far different ranges of \paco \cite{SLUTSKY19931833}.
However, decision split values for the two other influential features are \pfr and mean airway pressure—commonly used to assess the severity of acute respiratory distress seem to be more in line with clinical decision making at the bedside \cite{Villar2013-oi, Marini1992-zg}. 

Inspection of the terminal nodes shows that the RL policy selects distinct tidal volumes (ranging from approximately 3.4 to 8.3 mL/kg), PEEP values (5–12.5 cmH$_2$O), and FiO2 (0.4–0.5), depending on the interplay between CO2 levels, oxygenation (\pfr), and hemodynamic status (sysbp). In scenarios with very high mean airway pressure and moderately elevated PaCO2, the policy increases both tidal volume and PEEP. A plausible clinical rationale is that high PEEP may recruit alveoli when the lung mechanics are poor, whereas a larger tidal volume might aid in CO2 clearance \cite{Halter2007-sz}. Conversely, when mean airway pressure is low and \paco\text{ }is moderate, the policy opts for a markedly reduced tidal volume, suggesting an emphasis on lung-protective ventilation and prevention of overdistension. In the case of extremely high \paco\text{ }($\geq 92.1$ mmHg), the tree further partitions patients by systolic blood pressure. Higher PEEP (10 cmH$_2$O) is recommended only when systolic blood pressure is adequate ($\geq 96$ mmHg). This indicates a learned behavior that offsets the risk of compromised venous return and hypotension often associated with increased intrathoracic pressure \cite{Dorinsky1983-oa}. When blood pressure is lower, the tree prescribes moderate tidal volume (7.1 mL/kg) and minimal PEEP (5 cmH$_2$O), presumably to avoid hemodynamic deterioration.

The policy thus appears to balance hypercapnia and hypoxemia by adapting tidal volume and PEEP based on \paco\text{ }and \pfr, and it distinguishes heavier patients (above 105 kg) for potential differences in lung compliance or the need to approximate ideal body weight. Hemodynamic considerations also play a role when \paco\text{ }is extremely high, as indicated by a separate decision node for systolic blood pressure, demonstrating the RL model’s capacity to integrate respiratory support with cardiovascular stability. These patterns illustrate how interpretable RL policy representations can provide insight into data-driven ventilator adjustments that incorporate both oxygenation and ventilation goals, while accounting for body habitus and systemic perfusion pressures.

\section{Conclusion}

In conclusion, we present an interpretable RL policy for mechanical ventilation control and a novel causal non-parametric model-based approach for evaluating RL policies in an offline setting. By comparing various policies (behavior cloning, CQI and CQL), we empirically establish the superior performance of the latter two policies over the behavior clone to the clinicians as observed in MIMIC-III database in two ways: achieving greater increases in SpO$_2$ on average while taking aggressive actions less frequently, contributing to the broader goals of connected health by providing interpretable decision making systems that can integrate into ICU monitoring system and enhance clinical workflows.

However, it is crucial to acknowledge the inherent limitations of offline RL and offline off-policy evaluation (OPE) methods, which may suffer from issues such as data bias, distribution shift and lack of exploration. Although the model may outperform the clinicians for the goals set in this study, the goals set by the clinician are unknown as the clinical scenario is not included in the predictors, similarly to previous RL approaches to control ventilators. Therefore, while the results are promising, additional verification on more diverse datasets outside of MIMIC-III is necessary before clinical deployment. Nevertheless, the interpretable decision tree policies learned by CQI have the potential to provide useful clinical insights in future studies by highlighting the important features and decision rules. Moving forward, integrating more domain knowledge and considering safety constraints will be crucial steps towards developing trustworthy RL systems for safety-critical healthcare applications like mechanical ventilation control.

\begin{acks}
This material is based upon work supported by the National Science Foundation under Grant No. DGE-2125913 and Grant No. CMMI-1847666.
Image in figure \ref{fig:dynamics} is an image by pngtree.com.
\end{acks}

\bibliographystyle{ACM-Reference-Format}
\bibliography{refs}


\begin{thebibliography}{42}


\ifx \showCODEN    \undefined \def \showCODEN     #1{\unskip}     \fi
\ifx \showDOI      \undefined \def \showDOI       #1{#1}\fi
\ifx \showISBNx    \undefined \def \showISBNx     #1{\unskip}     \fi
\ifx \showISBNxiii \undefined \def \showISBNxiii  #1{\unskip}     \fi
\ifx \showISSN     \undefined \def \showISSN      #1{\unskip}     \fi
\ifx \showLCCN     \undefined \def \showLCCN      #1{\unskip}     \fi
\ifx \shownote     \undefined \def \shownote      #1{#1}          \fi
\ifx \showarticletitle \undefined \def \showarticletitle #1{#1}   \fi
\ifx \showURL      \undefined \def \showURL       {\relax}        \fi
\providecommand\bibfield[2]{#2}
\providecommand\bibinfo[2]{#2}
\providecommand\natexlab[1]{#1}
\providecommand\showeprint[2][]{arXiv:#2}

\bibitem[{Acute Respiratory Distress Syndrome Network} et~al\mbox{.}(2000)]%
        {lowtidalvolume}
\bibfield{author}{\bibinfo{person}{{Acute Respiratory Distress Syndrome Network}}, \bibinfo{person}{Roy~G Brower}, \bibinfo{person}{Michael~A Matthay}, \bibinfo{person}{Alan Morris}, \bibinfo{person}{David Schoenfeld}, \bibinfo{person}{B~Taylor Thompson}, {and} \bibinfo{person}{Arthur Wheeler}.} \bibinfo{year}{2000}\natexlab{}.
\newblock \showarticletitle{Ventilation with lower tidal volumes as compared with traditional tidal volumes for acute lung injury and the acute respiratory distress syndrome}.
\newblock \bibinfo{journal}{\emph{N Engl J Med}} \bibinfo{volume}{342}, \bibinfo{number}{18} (\bibinfo{date}{May} \bibinfo{year}{2000}), \bibinfo{pages}{1301--1308}.
\newblock


\bibitem[Bellani et~al\mbox{.}(2016)]%
        {Bellani2016}
\bibfield{author}{\bibinfo{person}{Giacomo Bellani} {et~al\mbox{.}}} \bibinfo{year}{2016}\natexlab{}.
\newblock \showarticletitle{Epidemiology, Patterns of Care, and Mortality for Patients With Acute Respiratory Distress Syndrome in Intensive Care Units in 50 Countries}.
\newblock \bibinfo{journal}{\emph{JAMA}} \bibinfo{volume}{315}, \bibinfo{number}{8} (\bibinfo{date}{Feb.} \bibinfo{year}{2016}), \bibinfo{pages}{788}.
\newblock
\showISSN{0098-7484}


\bibitem[Bennett and Kallus(2019)]%
        {NEURIPS2019_7c4bf50b}
\bibfield{author}{\bibinfo{person}{Andrew Bennett} {and} \bibinfo{person}{Nathan Kallus}.} \bibinfo{year}{2019}\natexlab{}.
\newblock \showarticletitle{Policy Evaluation with Latent Confounders via Optimal Balance}. In \bibinfo{booktitle}{\emph{Adv. Neural Inf. Process. Syst.}}, Vol.~\bibinfo{volume}{32}.
\newblock


\bibitem[Blauvelt et~al\mbox{.}(2022)]%
        {Blauvelt2022}
\bibfield{author}{\bibinfo{person}{David~G. Blauvelt} {et~al\mbox{.}}} \bibinfo{year}{2022}\natexlab{}.
\newblock \showarticletitle{Association of Ventilator Settings With Mortality in Pediatric Patients Treated With Extracorporeal Life Support for Respiratory Failure}.
\newblock \bibinfo{journal}{\emph{ASAIO Journal}} \bibinfo{volume}{68}, \bibinfo{number}{12} (\bibinfo{date}{March} \bibinfo{year}{2022}), \bibinfo{pages}{1536–1543}.
\newblock
\showISSN{1058-2916}


\bibitem[Britos et~al\mbox{.}(2011)]%
        {Britos2011}
\bibfield{author}{\bibinfo{person}{Martin Britos}, \bibinfo{person}{Elizabeth Smoot}, \bibinfo{person}{Kathleen~D. Liu}, \bibinfo{person}{B.~Taylor Thompson}, \bibinfo{person}{William Checkley}, {and} \bibinfo{person}{Roy~G. Brower}.} \bibinfo{year}{2011}\natexlab{}.
\newblock \showarticletitle{The value of positive end-expiratory pressure and Fio2 criteria in the definition of the acute respiratory distress syndrome*}.
\newblock \bibinfo{journal}{\emph{Critical Care Medicine}} \bibinfo{volume}{39}, \bibinfo{number}{9} (\bibinfo{date}{Sept.} \bibinfo{year}{2011}), \bibinfo{pages}{2025–2030}.
\newblock
\showISSN{0090-3493}


\bibitem[Brookhart et~al\mbox{.}(2010)]%
        {pmid20473199}
\bibfield{author}{\bibinfo{person}{M.~Alan Brookhart}, \bibinfo{person}{Til St\"{u}rmer}, \bibinfo{person}{Robert~J. Glynn}, \bibinfo{person}{Jeremy Rassen}, {and} \bibinfo{person}{Sebastian Schneeweiss}.} \bibinfo{year}{2010}\natexlab{}.
\newblock \showarticletitle{Confounding Control in Healthcare Database Research: Challenges and Potential Approaches}.
\newblock \bibinfo{journal}{\emph{Medical Care}} \bibinfo{volume}{48}, \bibinfo{number}{6} (\bibinfo{year}{2010}), \bibinfo{pages}{S114–S120}.
\newblock
\showISSN{0025-7079}


\bibitem[Curley et~al\mbox{.}(2016)]%
        {Curley2016}
\bibfield{author}{\bibinfo{person}{Gerard~F. Curley}, \bibinfo{person}{John~G. Laffey}, \bibinfo{person}{Haibo Zhang}, {and} \bibinfo{person}{Arthur~S. Slutsky}.} \bibinfo{year}{2016}\natexlab{}.
\newblock \showarticletitle{Biotrauma and Ventilator-Induced Lung Injury}.
\newblock \bibinfo{journal}{\emph{Chest}} \bibinfo{volume}{150}, \bibinfo{number}{5} (\bibinfo{date}{Nov.} \bibinfo{year}{2016}), \bibinfo{pages}{1109–1117}.
\newblock
\showISSN{0012-3692}


\bibitem[den Hengst et~al\mbox{.}(2024)]%
        {denHengst2024}
\bibfield{author}{\bibinfo{person}{Floris den Hengst}, \bibinfo{person}{Martijn Otten}, \bibinfo{person}{Paul Elbers}, \bibinfo{person}{Frank van Harmelen}, \bibinfo{person}{Vincent Fran\c{c}ois-Lavet}, {and} \bibinfo{person}{Mark Hoogendoorn}.} \bibinfo{year}{2024}\natexlab{}.
\newblock \showarticletitle{Guideline-informed reinforcement learning for mechanical ventilation in critical care}.
\newblock \bibinfo{journal}{\emph{Artificial Intelligence in Medicine}}  \bibinfo{volume}{147} (\bibinfo{date}{Jan.} \bibinfo{year}{2024}), \bibinfo{pages}{102742}.
\newblock
\showISSN{0933-3657}
\urldef\tempurl%
\url{https://doi.org/10.1016/j.artmed.2023.102742}
\showDOI{\tempurl}


\bibitem[Dorinsky and Whitcomb(1983)]%
        {Dorinsky1983-oa}
\bibfield{author}{\bibinfo{person}{Paul~M Dorinsky} {and} \bibinfo{person}{Michael~E Whitcomb}.} \bibinfo{year}{1983}\natexlab{}.
\newblock \showarticletitle{The effect of {PEEP} on cardiac output}.
\newblock \bibinfo{journal}{\emph{Chest}} \bibinfo{volume}{84}, \bibinfo{number}{2} (\bibinfo{date}{Aug.} \bibinfo{year}{1983}), \bibinfo{pages}{210--216}.
\newblock


\bibitem[Emerson et~al\mbox{.}(2023)]%
        {Emerson2023}
\bibfield{author}{\bibinfo{person}{Harry Emerson}, \bibinfo{person}{Matthew Guy}, {and} \bibinfo{person}{Ryan McConville}.} \bibinfo{year}{2023}\natexlab{}.
\newblock \showarticletitle{Offline reinforcement learning for safer blood glucose control in people with type 1 diabetes}.
\newblock \bibinfo{journal}{\emph{J. Biomed. Inform.}}  \bibinfo{volume}{142} (\bibinfo{date}{June} \bibinfo{year}{2023}), \bibinfo{pages}{104376}.
\newblock
\showISSN{1532-0464}
\urldef\tempurl%
\url{https://doi.org/10.1016/j.jbi.2023.104376}
\showDOI{\tempurl}


\bibitem[Ferguson et~al\mbox{.}(2012)]%
        {Ferguson2012}
\bibfield{author}{\bibinfo{person}{Niall~D. Ferguson} {et~al\mbox{.}}} \bibinfo{year}{2012}\natexlab{}.
\newblock \showarticletitle{The Berlin definition of ARDS: an expanded rationale, justification, and supplementary material}.
\newblock \bibinfo{journal}{\emph{Intensive Care Medicine}} \bibinfo{volume}{38}, \bibinfo{number}{10} (\bibinfo{date}{Aug.} \bibinfo{year}{2012}), \bibinfo{pages}{1573–1582}.
\newblock
\showISSN{1432-1238}
\urldef\tempurl%
\url{https://doi.org/10.1007/s00134-012-2682-1}
\showDOI{\tempurl}


\bibitem[Halter et~al\mbox{.}(2007)]%
        {Halter2007-sz}
\bibfield{author}{\bibinfo{person}{Jeffrey~M Halter}, \bibinfo{person}{Jay~M Steinberg}, \bibinfo{person}{Louis~A Gatto}, \bibinfo{person}{Joseph~D DiRocco}, \bibinfo{person}{Lucio~A Pavone}, \bibinfo{person}{Henry~J Schiller}, \bibinfo{person}{Scott Albert}, \bibinfo{person}{Hsi-Ming Lee}, \bibinfo{person}{David Carney}, {and} \bibinfo{person}{Gary~F Nieman}.} \bibinfo{year}{2007}\natexlab{}.
\newblock \showarticletitle{Effect of positive end-expiratory pressure and tidal volume on lung injury induced by alveolar instability}.
\newblock \bibinfo{journal}{\emph{Crit. Care}} \bibinfo{volume}{11}, \bibinfo{number}{1} (\bibinfo{year}{2007}), \bibinfo{pages}{R20}.
\newblock


\bibitem[Johnson et~al\mbox{.}(2016)]%
        {Johnson2016}
\bibfield{author}{\bibinfo{person}{Alistair~E.W. Johnson} {et~al\mbox{.}}} \bibinfo{year}{2016}\natexlab{}.
\newblock \showarticletitle{{MIMIC-III}, a freely accessible critical care database}.
\newblock \bibinfo{journal}{\emph{Scientific Data}} \bibinfo{volume}{3}, \bibinfo{number}{1} (\bibinfo{year}{2016}).
\newblock
\showISSN{2052-4463}


\bibitem[Komorowski et~al\mbox{.}(2018)]%
        {KCB18}
\bibfield{author}{\bibinfo{person}{Matthieu Komorowski}, \bibinfo{person}{Leo~A. Celi}, \bibinfo{person}{Omar Badawi}, \bibinfo{person}{Anthony~C. Gordon}, {and} \bibinfo{person}{A.~Aldo Faisal}.} \bibinfo{year}{2018}\natexlab{}.
\newblock \showarticletitle{The Artificial Intelligence Clinician learns optimal treatment strategies for sepsis in intensive care}.
\newblock \bibinfo{journal}{\emph{Nature Medicine}} \bibinfo{volume}{24}, \bibinfo{number}{11} (\bibinfo{date}{01 Nov} \bibinfo{year}{2018}), \bibinfo{pages}{1716--1720}.
\newblock
\showISSN{1546-170X}
\urldef\tempurl%
\url{https://doi.org/10.1038/s41591-018-0213-5}
\showDOI{\tempurl}


\bibitem[Kondrup et~al\mbox{.}(2023)]%
        {deepvent}
\bibfield{author}{\bibinfo{person}{Flemming Kondrup} {et~al\mbox{.}}} \bibinfo{year}{2023}\natexlab{}.
\newblock \showarticletitle{Towards Safe Mechanical Ventilation Treatment Using Deep Offline Reinforcement Learning}. In \bibinfo{booktitle}{\emph{Proc. AAAI Conf. Artif. Intell.}}, Vol.~\bibinfo{volume}{37}. \bibinfo{pages}{15696--15702}.
\newblock
\urldef\tempurl%
\url{https://doi.org/10.1609/aaai.v37i13.26862}
\showDOI{\tempurl}


\bibitem[Kumar et~al\mbox{.}(2020)]%
        {cql}
\bibfield{author}{\bibinfo{person}{Aviral Kumar}, \bibinfo{person}{Aurick Zhou}, \bibinfo{person}{George Tucker}, {and} \bibinfo{person}{Sergey Levine}.} \bibinfo{year}{2020}\natexlab{}.
\newblock \showarticletitle{Conservative {Q-L}earning for Offline Reinforcement Learning}. In \bibinfo{booktitle}{\emph{Adv. Neural Inf. Process. Syst.}}, Vol.~\bibinfo{volume}{33}. \bibinfo{pages}{1179--1191}.
\newblock


\bibitem[Le et~al\mbox{.}(2019)]%
        {fqe}
\bibfield{author}{\bibinfo{person}{Hoang Le}, \bibinfo{person}{Cameron Voloshin}, {and} \bibinfo{person}{Yisong Yue}.} \bibinfo{year}{2019}\natexlab{}.
\newblock \showarticletitle{Batch Policy Learning under Constraints}. In \bibinfo{booktitle}{\emph{Proceedings of the 36th International Conference on Machine Learning}} \emph{(\bibinfo{series}{Proceedings of Machine Learning Research}, Vol.~\bibinfo{volume}{97})}, \bibfield{editor}{\bibinfo{person}{Kamalika Chaudhuri} {and} \bibinfo{person}{Ruslan Salakhutdinov}} (Eds.). \bibinfo{publisher}{PMLR}, \bibinfo{pages}{3703--3712}.
\newblock


\bibitem[Lee et~al\mbox{.}(2023)]%
        {Lee2023}
\bibfield{author}{\bibinfo{person}{Hyeonhoon Lee}, \bibinfo{person}{Hyun-Kyu Yoon}, \bibinfo{person}{Jaewon Kim}, \bibinfo{person}{Ji~Soo Park}, \bibinfo{person}{Chang-Hoon Koo}, \bibinfo{person}{Dongwook Won}, {and} \bibinfo{person}{Hyung-Chul Lee}.} \bibinfo{year}{2023}\natexlab{}.
\newblock \showarticletitle{Development and validation of a reinforcement learning model for ventilation control during emergence from general anesthesia}.
\newblock \bibinfo{journal}{\emph{npj Digital Medicine}} \bibinfo{volume}{6}, \bibinfo{number}{1} (\bibinfo{date}{Aug.} \bibinfo{year}{2023}).
\newblock
\showISSN{2398-6352}
\urldef\tempurl%
\url{https://doi.org/10.1038/s41746-023-00893-w}
\showDOI{\tempurl}


\bibitem[Liu et~al\mbox{.}(2017)]%
        {Liu2017}
\bibfield{author}{\bibinfo{person}{Ying Liu}, \bibinfo{person}{Brent Logan}, \bibinfo{person}{Ning Liu}, \bibinfo{person}{Zhiyuan Xu}, \bibinfo{person}{Jian Tang}, {and} \bibinfo{person}{Yangzhi Wang}.} \bibinfo{year}{2017}\natexlab{}.
\newblock \showarticletitle{Deep Reinforcement Learning for Dynamic Treatment Regimes on Medical Registry Data}. In \bibinfo{booktitle}{\emph{2017 IEEE International Conference on Healthcare Informatics (ICHI)}}. \bibinfo{publisher}{IEEE}, \bibinfo{pages}{380–385}.
\newblock
\urldef\tempurl%
\url{https://doi.org/10.1109/ichi.2017.45}
\showDOI{\tempurl}


\bibitem[Loss et~al\mbox{.}(2015)]%
        {Loss2015-om}
\bibfield{author}{\bibinfo{person}{S{\'e}rgio~Henrique Loss} {et~al\mbox{.}}} \bibinfo{year}{2015}\natexlab{}.
\newblock \showarticletitle{The reality of patients requiring prolonged mechanical ventilation: a multicenter study}.
\newblock \bibinfo{journal}{\emph{Rev. Bras. Ter. Intensiva}} \bibinfo{volume}{27}, \bibinfo{number}{1} (\bibinfo{year}{2015}), \bibinfo{pages}{26--35}.
\newblock


\bibitem[Luo et~al\mbox{.}(2024)]%
        {pmlr-v235-luo24f}
\bibfield{author}{\bibinfo{person}{Zhiyao Luo}, \bibinfo{person}{Yangchen Pan}, \bibinfo{person}{Peter Watkinson}, {and} \bibinfo{person}{Tingting Zhu}.} \bibinfo{year}{2024}\natexlab{}.
\newblock \showarticletitle{Position: Reinforcement Learning in Dynamic Treatment Regimes Needs Critical Reexamination}. In \bibinfo{booktitle}{\emph{Proceedings of the 41st International Conference on Machine Learning}} \emph{(\bibinfo{series}{Proceedings of Machine Learning Research}, Vol.~\bibinfo{volume}{235})}, \bibfield{editor}{\bibinfo{person}{Ruslan Salakhutdinov}, \bibinfo{person}{Zico Kolter}, \bibinfo{person}{Katherine Heller}, \bibinfo{person}{Adrian Weller}, \bibinfo{person}{Nuria Oliver}, \bibinfo{person}{Jonathan Scarlett}, {and} \bibinfo{person}{Felix Berkenkamp}} (Eds.). \bibinfo{publisher}{PMLR}, \bibinfo{pages}{33432--33465}.
\newblock


\bibitem[Marini and Ravenscraft(1992)]%
        {Marini1992-zg}
\bibfield{author}{\bibinfo{person}{J~J Marini} {and} \bibinfo{person}{S~A Ravenscraft}.} \bibinfo{year}{1992}\natexlab{}.
\newblock \showarticletitle{Mean airway pressure: physiologic determinants and clinical importance--Part 2: Clinical implications}.
\newblock \bibinfo{journal}{\emph{Crit. Care Med.}} \bibinfo{volume}{20}, \bibinfo{number}{11} (\bibinfo{date}{Nov.} \bibinfo{year}{1992}), \bibinfo{pages}{1604--1616}.
\newblock


\bibitem[Nadaraya(1964)]%
        {Nadaraya1964OnER}
\bibfield{author}{\bibinfo{person}{Elizbar Nadaraya}.} \bibinfo{year}{1964}\natexlab{}.
\newblock \showarticletitle{On Estimating Regression}.
\newblock \bibinfo{journal}{\emph{Theory of Probability and Its Applications}}  \bibinfo{volume}{9} (\bibinfo{year}{1964}), \bibinfo{pages}{141--142}.
\newblock


\bibitem[Nanayakkara et~al\mbox{.}(2022)]%
        {NCLS22}
\bibfield{author}{\bibinfo{person}{Thesath Nanayakkara}, \bibinfo{person}{Gilles Clermont}, \bibinfo{person}{Christopher~James Langmead}, {and} \bibinfo{person}{David Swigon}.} \bibinfo{year}{2022}\natexlab{}.
\newblock \showarticletitle{Unifying cardiovascular modelling with deep reinforcement learning for uncertainty aware control of sepsis treatment}.
\newblock \bibinfo{journal}{\emph{PLOS Digital Health}} \bibinfo{volume}{1}, \bibinfo{number}{2} (\bibinfo{date}{02} \bibinfo{year}{2022}), \bibinfo{pages}{1--20}.
\newblock


\bibitem[Oberst and Sontag(2019)]%
        {pmlr-v97-oberst19a}
\bibfield{author}{\bibinfo{person}{Michael Oberst} {and} \bibinfo{person}{David Sontag}.} \bibinfo{year}{2019}\natexlab{}.
\newblock \showarticletitle{Counterfactual Off-Policy Evaluation with {G}umbel-Max Structural Causal Models}. In \bibinfo{booktitle}{\emph{Proc. Int. Conf. Mach. Learn.}}, Vol.~\bibinfo{volume}{97}. \bibinfo{pages}{4881--4890}.
\newblock


\bibitem[Oroojeni Mohammad~Javad et~al\mbox{.}(2019)]%
        {Oroojeni_Mohammad_Javad2019-lw}
\bibfield{author}{\bibinfo{person}{Mahsa Oroojeni Mohammad~Javad}, \bibinfo{person}{Stephen~Olusegun Agboola}, \bibinfo{person}{Kamal Jethwani}, \bibinfo{person}{Abe Zeid}, {and} \bibinfo{person}{Sagar Kamarthi}.} \bibinfo{year}{2019}\natexlab{}.
\newblock \showarticletitle{A Reinforcement Learning–Based Method for Management of Type 1 Diabetes: Exploratory Study}.
\newblock \bibinfo{journal}{\emph{JMIR Diabetes}} \bibinfo{volume}{4}, \bibinfo{number}{3} (\bibinfo{date}{Aug.} \bibinfo{year}{2019}), \bibinfo{pages}{e12905}.
\newblock
\showISSN{2371-4379}
\urldef\tempurl%
\url{https://doi.org/10.2196/12905}
\showDOI{\tempurl}


\bibitem[O’Brien et~al\mbox{.}(2007)]%
        {OBRIEN20071012}
\bibfield{author}{\bibinfo{person}{James~M. O’Brien}, \bibinfo{person}{Naeem~A. Ali}, \bibinfo{person}{Scott~K. Aberegg}, {and} \bibinfo{person}{Edward Abraham}.} \bibinfo{year}{2007}\natexlab{}.
\newblock \showarticletitle{Sepsis}.
\newblock \bibinfo{journal}{\emph{The American Journal of Medicine}} \bibinfo{volume}{120}, \bibinfo{number}{12} (\bibinfo{year}{2007}), \bibinfo{pages}{1012--1022}.
\newblock
\showISSN{0002-9343}
\urldef\tempurl%
\url{https://doi.org/10.1016/j.amjmed.2007.01.035}
\showDOI{\tempurl}


\bibitem[Parbhoo et~al\mbox{.}(2017)]%
        {Parbhoo2017-zr}
\bibfield{author}{\bibinfo{person}{Sonali Parbhoo}, \bibinfo{person}{Jasmina Bogojeska}, \bibinfo{person}{Maurizio Zazzi}, \bibinfo{person}{Volker Roth}, {and} \bibinfo{person}{Finale Doshi-Velez}.} \bibinfo{year}{2017}\natexlab{}.
\newblock \showarticletitle{Combining Kernel and Model Based Learning for {HIV} Therapy Selection}.
\newblock \bibinfo{journal}{\emph{AMIA Jt Summits Transl. Sci. Proc.}}  \bibinfo{volume}{2017} (\bibinfo{date}{July} \bibinfo{year}{2017}), \bibinfo{pages}{239--248}.
\newblock


\bibitem[Peine et~al\mbox{.}(2021)]%
        {ventai}
\bibfield{author}{\bibinfo{person}{Arne Peine} {et~al\mbox{.}}} \bibinfo{year}{2021}\natexlab{}.
\newblock \showarticletitle{Development and validation of a reinforcement learning algorithm to dynamically optimize mechanical ventilation in critical care}.
\newblock \bibinfo{journal}{\emph{npj Digital Medicine}} \bibinfo{volume}{4}, \bibinfo{number}{1} (\bibinfo{year}{2021}), \bibinfo{pages}{32}.
\newblock
\showISBNx{2398-6352}
\urldef\tempurl%
\url{https://doi.org/10.1038/s41746-021-00388-6}
\showDOI{\tempurl}


\bibitem[Pinheiro~de Oliveira et~al\mbox{.}(2010)]%
        {PHADF2010}
\bibfield{author}{\bibinfo{person}{Roselaine Pinheiro~de Oliveira}, \bibinfo{person}{Marcio~Pereira Hetzel}, \bibinfo{person}{Mauro dos Anjos~Silva}, \bibinfo{person}{Daniele Dallegrave}, {and} \bibinfo{person}{Gilberto Friedman}.} \bibinfo{year}{2010}\natexlab{}.
\newblock \showarticletitle{Mechanical ventilation with high tidal volume induces inflammation in patients without lung disease}.
\newblock \bibinfo{journal}{\emph{Critical Care}} \bibinfo{volume}{14}, \bibinfo{number}{2} (\bibinfo{date}{March} \bibinfo{year}{2010}).
\newblock
\showISSN{1364-8535}


\bibitem[Prasad et~al\mbox{.}(2017)]%
        {prasad2017reinforcement}
\bibfield{author}{\bibinfo{person}{Niranjani Prasad}, \bibinfo{person}{Li~Fang Cheng}, \bibinfo{person}{Corey Chivers}, \bibinfo{person}{Michael Draugelis}, {and} \bibinfo{person}{Barbara~E Engelhardt}.} \bibinfo{year}{2017}\natexlab{}.
\newblock \showarticletitle{A reinforcement learning approach to weaning of mechanical ventilation in intensive care units}. In \bibinfo{booktitle}{\emph{33rd Conference on Uncertainty in Artificial Intelligence, UAI 2017}}.
\newblock


\bibitem[Precup et~al\mbox{.}(2000)]%
        {precup_is}
\bibfield{author}{\bibinfo{person}{Doina Precup}, \bibinfo{person}{Richard~S. Sutton}, {and} \bibinfo{person}{Satinder~P. Singh}.} \bibinfo{year}{2000}\natexlab{}.
\newblock \showarticletitle{Eligibility Traces for Off-Policy Policy Evaluation}. In \bibinfo{booktitle}{\emph{Proceedings of the Seventeenth International Conference on Machine Learning}} \emph{(\bibinfo{series}{ICML '00})}. \bibinfo{publisher}{Morgan Kaufmann Publishers Inc.}, \bibinfo{address}{San Francisco, CA, USA}, \bibinfo{pages}{759–766}.
\newblock
\showISBNx{1558607072}


\bibitem[Rachmale et~al\mbox{.}(2012)]%
        {Rachmale1887}
\bibfield{author}{\bibinfo{person}{Sonal Rachmale}, \bibinfo{person}{Guangxi Li}, \bibinfo{person}{Gregory Wilson}, \bibinfo{person}{Michael Malinchoc}, {and} \bibinfo{person}{Ognjen Gajic}.} \bibinfo{year}{2012}\natexlab{}.
\newblock \showarticletitle{Practice of Excessive FIO2and Effect on Pulmonary Outcomes in Mechanically Ventilated Patients With Acute Lung Injury}.
\newblock \bibinfo{journal}{\emph{Respiratory Care}} \bibinfo{volume}{57}, \bibinfo{number}{11} (\bibinfo{date}{Nov.} \bibinfo{year}{2012}), \bibinfo{pages}{1887–1893}.
\newblock
\showISSN{1943-3654}


\bibitem[Roth et~al\mbox{.}(2019)]%
        {cqi}
\bibfield{author}{\bibinfo{person}{Aaron~M. Roth}, \bibinfo{person}{Nicholay Topin}, \bibinfo{person}{Pooyan Jamshidi}, {and} \bibinfo{person}{Manuela Veloso}.} \bibinfo{year}{2019}\natexlab{}.
\newblock \showarticletitle{Conservative {Q-I}mprovement: Reinforcement Learning for an Interpretable Decision-Tree Policy}.
\newblock \bibinfo{journal}{\emph{arXiv:1907.01180}} (\bibinfo{year}{2019}).
\newblock


\bibitem[Schnapp and Cohen(1990)]%
        {spo23}
\bibfield{author}{\bibinfo{person}{Lynn~M. Schnapp} {and} \bibinfo{person}{Neal~H. Cohen}.} \bibinfo{year}{1990}\natexlab{}.
\newblock \showarticletitle{Pulse Oximetry: Uses and Abuses}.
\newblock \bibinfo{journal}{\emph{Chest}} \bibinfo{volume}{98}, \bibinfo{number}{5} (\bibinfo{year}{1990}), \bibinfo{pages}{1244--1250}.
\newblock
\showISSN{0012-3692}


\bibitem[Seno and Imai(2022)]%
        {d3rlpy}
\bibfield{author}{\bibinfo{person}{Takuma Seno} {and} \bibinfo{person}{Michita Imai}.} \bibinfo{year}{2022}\natexlab{}.
\newblock \showarticletitle{d3rlpy: An Offline Deep Reinforcement Learning Library}.
\newblock \bibinfo{journal}{\emph{J. Mach. Learn. Res.}}  \bibinfo{volume}{23} (\bibinfo{date}{1 Oct.} \bibinfo{year}{2022}).
\newblock
\showISSN{1532-4435}


\bibitem[Slutsky(1993)]%
        {SLUTSKY19931833}
\bibfield{author}{\bibinfo{person}{Arthur~S. Slutsky}.} \bibinfo{year}{1993}\natexlab{}.
\newblock \showarticletitle{Mechanical Ventilation}.
\newblock \bibinfo{journal}{\emph{Chest}} \bibinfo{volume}{104}, \bibinfo{number}{6} (\bibinfo{year}{1993}), \bibinfo{pages}{1833--1859}.
\newblock
\showISSN{0012-3692}
\urldef\tempurl%
\url{https://doi.org/10.1378/chest.104.6.1833}
\showDOI{\tempurl}


\bibitem[Slutsky and Ranieri(2013)]%
        {SR2013}
\bibfield{author}{\bibinfo{person}{Arthur~S. Slutsky} {and} \bibinfo{person}{V.~Marco Ranieri}.} \bibinfo{year}{2013}\natexlab{}.
\newblock \showarticletitle{Ventilator-Induced Lung Injury}.
\newblock \bibinfo{journal}{\emph{New England Journal of Medicine}} \bibinfo{volume}{369}, \bibinfo{number}{22} (\bibinfo{date}{Nov.} \bibinfo{year}{2013}), \bibinfo{pages}{2126–2136}.
\newblock
\showISSN{1533-4406}


\bibitem[{van den Boom} et~al\mbox{.}(2020)]%
        {VANDENBOOM2020566}
\bibfield{author}{\bibinfo{person}{Willem {van den Boom}}, \bibinfo{person}{Michael Hoy}, \bibinfo{person}{Jagadish Sankaran}, \bibinfo{person}{Mengru Liu}, \bibinfo{person}{Haroun Chahed}, \bibinfo{person}{Mengling Feng}, {and} \bibinfo{person}{Kay~Choong See}.} \bibinfo{year}{2020}\natexlab{}.
\newblock \showarticletitle{The Search for Optimal Oxygen Saturation Targets in Critically Ill Patients: Observational Data From Large ICU Databases}.
\newblock \bibinfo{journal}{\emph{Chest}} \bibinfo{volume}{157}, \bibinfo{number}{3} (\bibinfo{year}{2020}), \bibinfo{pages}{566--573}.
\newblock
\showISSN{0012-3692}
\urldef\tempurl%
\url{https://doi.org/10.1016/j.chest.2019.09.015}
\showDOI{\tempurl}


\bibitem[Villar et~al\mbox{.}(2013)]%
        {Villar2013-oi}
\bibfield{author}{\bibinfo{person}{Jes{\'u}s Villar}, \bibinfo{person}{Lina P{\'e}rez-M{\'e}ndez}, \bibinfo{person}{Jes{\'u}s Blanco}, \bibinfo{person}{Jos{\'e}~Manuel A{\~n}{\'o}n}, \bibinfo{person}{Llu{\'\i}s Blanch}, \bibinfo{person}{Javier Belda}, \bibinfo{person}{Antonio Santos-Bouza}, \bibinfo{person}{Rosa~Lidia Fern{\'a}ndez}, \bibinfo{person}{Robert~M Kacmarek}, {and} \bibinfo{person}{{Spanish Initiative for Epidemiology, Stratification, and Therapies for ARDS (SIESTA) Network}}.} \bibinfo{year}{2013}\natexlab{}.
\newblock \showarticletitle{A universal definition of {ARDS}: the {PaO2/FiO2} ratio under a standard ventilatory setting--a prospective, multicenter validation study}.
\newblock \bibinfo{journal}{\emph{Intensive Care Med.}} \bibinfo{volume}{39}, \bibinfo{number}{4} (\bibinfo{date}{April} \bibinfo{year}{2013}), \bibinfo{pages}{583--592}.
\newblock


\bibitem[Watson(1964)]%
        {watson}
\bibfield{author}{\bibinfo{person}{Geoffrey~S. Watson}.} \bibinfo{year}{1964}\natexlab{}.
\newblock \showarticletitle{Smooth Regression Analysis}.
\newblock \bibinfo{journal}{\emph{Sankhyā: The Indian J. Statist., Series A (1961-2002)}} \bibinfo{volume}{26}, \bibinfo{number}{4} (\bibinfo{year}{1964}), \bibinfo{pages}{359--372}.
\newblock
\showISSN{0581572X}


\bibitem[Yu et~al\mbox{.}(2019)]%
        {Yu2019}
\bibfield{author}{\bibinfo{person}{Chao Yu}, \bibinfo{person}{Yinzhao Dong}, \bibinfo{person}{Jiming Liu}, {and} \bibinfo{person}{Guoqi Ren}.} \bibinfo{year}{2019}\natexlab{}.
\newblock \showarticletitle{Incorporating causal factors into reinforcement learning for dynamic treatment regimes in {HIV}}.
\newblock \bibinfo{journal}{\emph{BMC Med. Inform. and Decis. Making}} \bibinfo{volume}{19}, \bibinfo{number}{2} (\bibinfo{date}{09 Apr} \bibinfo{year}{2019}), \bibinfo{pages}{60}.
\newblock
\showISSN{1472-6947}


\end{thebibliography}

\appendix

\section{Research Methods}
\subsection{List of ICD-9 codes relevant to sepsis}
\begin{figure}[H]
    \centering
    \begin{tabular}{|c|c|c|c|c|c|}
        \hline
        \multicolumn{6}{|c|}{\textbf{ICD-9 Codes}} \\
        \hline
        024 & 036.3 & 038.0 & 038.2 & 038.3 & 038.40 \\
        038.41 & 038.42 & 038.43 & 038.44 & 038.49 & 038.8 \\
        038.9 & 098.89 & 360.00 & 519.01 & 522.4 & 528.3 \\
        567.22 & 599.0 & 614.9 & 630 & 632 & 634 \\
        635 & 636 & 637 & 638 & 639 & 659.3 \\
        670.2 & 670.3 & 672.00 & 682.9 & 771.3 & 771.81 \\
        771.89 & 785.52 & 791 & 995.91 & 995.92 & 995.94 \\
        996.64 & 998.02 & 998.59 & 999.31 & 999.39 & \\
        \hline
    \end{tabular}
    \label{fig:icd9codes}
    \Description[<short description>]{<long description>}
\end{figure}

\subsection{Action Binning Interval}

\begin{figure}[!ht]
    \small
    \centering
    \begin{tabular}{|c|c|c|c|}
        \hline
        bin & Vt$_{\text{set}}$ & PEEP & FiO$_2$ \\
        \hline
        1 & [0, 3.9) & [0, 7) & [0, 0.36)\\
        2 & [3.9, 5.37) & [7, 11) & [0.36, 0.45) \\
        3 & [5.37, 6.55) & [11, 16) & [0.45, 0.55)\\
        4 & [6.55, 7.74) & [16, $\infty$) & [0.55, 0.65)\\
        5 & [7.74, 9.12) & & [0.65, 0.76)\\ 
        6 & [9.12, 11.11) & & [0.76, 0.89) \\
        7 & [11.11, $\infty$) & & [0.89, 1]\\
        \hline
    \end{tabular}
    \caption{Action space binning intervals}
    \label{fig:action-bins}
    \Description[<short description>]{<long description>}
\end{figure}
Since the action space has been discretized, the action indices for Vt$_{\text{set}}$ and FiO$_{2}$ including and above 6 and 4 respectively, out of 1-7, are considered aggressive, as they each contain $10$ and $0.6$.

\subsection{Transition Model Feature grouping}
\begin{enumerate}
    \item (Respiratory): Respiratory rate, Spontaneous tidal volume, \pfr, Mean airway pressure
    \item (Hemodynamic): Heart rate, Systolic BP, Diastolic BP
    \item (Blood Gas): \spo, \paco, \pao
    \item (Miscellaneous): Sepsis, Weight, Age, GCS, Cumulative Fluid Balance
\end{enumerate}

\subsection{Transition Model Hyperparameters explored}
\begin{enumerate}
    \item $h_{s,h}$: $[2.036, 2.536, 3.036]$
    \item $h_{s,r}$: $[1.8, 2.3, 2.8]$
    \item $h_{s,b}$: $[1.532, 2.032, 2.532]$
    \item $h_{s,m}$: $[1.532, 2.032, 2.532]$
    \item $h_a$: $[1.0, 1.5, 2.0]$
    \item $h_z$: $[0.5, 1.0, 1.5, 2.0]$
    \item $\lambda$: $[10^{-2}, 10^{-3}, 10^{-4}]$
\end{enumerate}

\end{document}